%% file: main.tex
\documentclass{article}
\usepackage[final, nonatbib]{neurips_2020}

\usepackage[utf8]{inputenc} 
\usepackage[T1]{fontenc}    
\usepackage[colorlinks=true,linkcolor=blue,citecolor=blue]{hyperref}       
\usepackage[round]{natbib}
\usepackage{url}            
\usepackage{booktabs}       
\usepackage{amsfonts}       
\usepackage{nicefrac}       
\usepackage{microtype}      
\usepackage{physics}
\usepackage{wrapfig}
\usepackage[toc,page,header]{appendix}
\usepackage{minitoc}
\usepackage{amsmath}
\usepackage{amssymb}
\usepackage{amsthm}
\usepackage{amsfonts}
\usepackage{mathtools}
\usepackage[bb=boondox,bbscaled=.95]{mathalfa}
\usepackage{bm}
\usepackage{subfigure}
\usepackage{cancel}
\usepackage{thmtools, thm-restate}
\usepackage{hyperref}
\usepackage[most]{tcolorbox}
\usepackage{pgfplots}
\usepackage{tikz}
\usepackage[frozencache, cachedir=./]{minted} %

\usepgfplotslibrary{groupplots}
\usepgfplotslibrary{patchplots}
\pgfplotsset{compat=1.16}
\declaretheorem{theorem}

\newtheorem{rmk}{Remark}

\newcommand{\R}{\mathbb{R}}
\newcommand{\N}{\mathbb{N}}

\newcommand{\bL}{\mathbb{L}}

\newcommand{\cO}{\mathcal{O}}

\newcommand{\cS}{\mathcal{S}}
\newcommand{\cL}{\mathcal{L}}

\newcommand{\cW}{\mathcal{W}}
\newcommand{\cK}{\mathcal{K}}

\newcommand{\nz}{{n_z}}
\newcommand{\nx}{{n_x}}
\newcommand{\ny}{{n_y}}
\newcommand{\nt}{{n_\theta}}
\newcommand{\nw}{{n_\omega}}

\newcommand{\ra}{\rightarrow}

\newcommand{\z}{\mathbf{z}}
\newcommand{\xb}{\mathbf{x}}
\newcommand{\wb}{\mathbf{w}}
\newcommand{\y}{\mathbf{y}}
\newcommand{\ab}{\mathbf{a}}

\definecolor{olive}{rgb}{0.6, 0.6, 0.2}
\definecolor{sand}{rgb}{0.8666666666666667, 0.8, 0.4666666666666667}
\definecolor{wine}{rgb}{0.5333333333333333, 0.13333333333333333, 0.3333333333333333}
\definecolor{deblue}{RGB}{11,132,147}
\definecolor{ocra}{RGB}{204, 119, 34}

\newcommand{\fcircle}[2][red,fill=red]{\tikz[baseline=-0.5ex]\draw[#1,radius=#2] (0,0.03) circle ;}
\newtcolorbox{CatchyBox}[2][]{
    lower separated=false,
    colback=white!80!sand!90!ocra,
    colframe=white, fonttitle=\bfseries,
    colbacktitle=white!50!sand!90!ocra,
    coltitle=black,
    enhanced,
    attach boxed title to top left={xshift=.02\linewidth,yshift=-4mm},
    title=#2,#1}

\title{Dissecting Neural ODEs}
\author{\ Stefano Massaroli$^*$\\
	\normalsize The University of Tokyo, {\tt DiffEqML}\\
	\fontsize{9}{10}\selectfont{\texttt{massaroli@robot.t.u-tokyo.ac.jp}}
	\And \normalsize
	Michael Poli\thanks{Equal contribution. Author order was decided by flipping a coin.}\\
	\normalsize KAIST, {\tt DiffEqML}\\
	\fontsize{9}{10}\selectfont{\texttt{poli\_m@kaist.ac.kr}}
	\AND \normalsize
    Jinkyoo Park\\
\normalsize	KAIST\\
	\fontsize{8}{9}\selectfont{\texttt{jinkyoo.park@kaist.ac.kr}}
	\And \normalsize
	Atsushi Yamashita\\
\normalsize	The University of Tokyo\\
	\fontsize{8}{9}\selectfont{\texttt{yamashita@robot.t.u-tokyo.ac.jp}}
	\And 
\normalsize	Hajime Asama \\c
\normalsize	The University of Tokyo\\
	\fontsize{8}{9}\selectfont{\texttt{asama@robot.t.u-tokyo.ac.jp}}
}

\begin{document}
\maketitle
\vspace*{-5mm}
\begin{abstract}
\vspace*{-3mm}
Continuous deep learning architectures have recently re--emerged as \textit{Neural Ordinary Differential Equations} (Neural ODEs). This infinite--depth approach theoretically bridges the gap between deep learning and dynamical systems, offering a novel perspective. However, deciphering the inner working of these models is still an open challenge, as most applications apply them as generic \textit{black--box} modules. In this work we ``open the box'', further developing the continuous--depth formulation with the aim of clarifying the influence of several design choices on the underlying dynamics. 
\end{abstract}
\doparttoc
\faketableofcontents
%
\input{1_Introduction}
\input{2_Background}
\input{3_NeuralDEs}
\input{4_Augmentation}
\input{5_Non_Augmentation}
\input{6_Related_Work}
\input{7_Discussion}
\section*{Broader Impact}
As continuous deep learning sees increased utilization across fields such as healthcare \citep{rubanova2019latent,yildiz2019ode}, it is of utmost importance that we develop appropriate tools to further our understanding of neural differential equations. The search for robustness in traditional deep learning has only recently seen a surge in ideas and proposed solutions; this work aims at providing exploratory first steps necessary to extend the discussion to this novel paradigm. The \textit{leitmotif} of this work is injecting system--theoretic concepts into the framework of continuous models. These ideas are of foundational importance in tangential fields such control and forecasting of dynamical systems, and are routinely used to develop robust algorithms with theoretical and practical guarantees. 
\section*{Acknowledgment}
This work was supported by the Basic Science Research Program through the National Research Foundation of Korea (NRF) funded by the Ministry of Education, 2018R1D1A1B07050443.
\bibliographystyle{abbrvnat}
\bibliography{main.bib}
%
%
\newpage
\rule[0pt]{\columnwidth}{3pt}
\begin{center}
\huge{\bf{Dissecting Neural ODEs} \\
\emph{Supplementary Material}}
\end{center}
\vspace*{3mm}
\rule[0pt]{\columnwidth}{1pt}
\vspace*{-.5in}
\vspace*{1in}

\appendix
\addcontentsline{toc}{section}{}
\part{}
\parttoc
\input{Appendix_A}
\input{Appendix_B}
%
\input{./Appendix_C.tex}
\end{document}

%% file: 1_Introduction.tex
\section{Introduction}
Neural ODEs \citep{chen2018neural} represent the latest instance of
continuous deep learning models, first developed in the context of continuous recurrent networks \citep{cohen1983absolute}. Since their introduction, research on Neural ODEs variants \citep{tzen2019neural,jia2019neural,zhang2019anodev2,yildiz2019ode,poli2019graph} has progressed at a rapid pace. However, the search for concise explanations and experimental evaluations of novel architectures has left many fundamental questions unanswered.

In this work, we establish a general system--theoretic Neural ODE formulation (\ref{eq:1}) and dissect it into its core components; we analyze each of them separately, shining light on peculiar phenomena unique to the continuous deep learning paradigm. In particular, augmentation strategies are generalized beyond ANODEs \citep{dupont2019augmented}, and the novel concepts of \textit{data--control} and \textit{adaptive--depth} enriching (\ref{eq:1}) are showcased as effective approaches to learn maps such as \textit{reflections} or \textit{concentric annuli} without augmentation.

While explicit dependence on the depth--variable has been considered in the original formulation \citep{chen2018neural}, a parameter depth--variance in continuous models has been overlooked. We provide a treatment in infinite--dimensional space required by the true \textit{deep limit} of ResNets, the solution of which leads to a Neural ODE variant based on a spectral discretization.
\begin{CatchyBox}{Neural Ordinary Differential Equation}
    \begin{minipage}[h]{0.45\linewidth}
            \begin{equation}\label{eq:1}
            {
                \left\{
                \begin{aligned}
                    \dot{\z}(s) & = f_{\color{wine}\bm{\theta(s)}}(s,{\color{olive} \xb}, \z(s))\\
                    \z(0) &= {\color{deblue}\bm{ h_x({\xb})}}\\
                    \hat{\y}(s) &=h_y(\z(s))
                \end{aligned}
                \right.~~
                s\in {\color{olive}\bm\cS}}
            \end{equation}
            \hfill
    \end{minipage}
    \hfill
    \begin{minipage}[h]{.51\linewidth}\small
        \centering
        \begin{tabular}{r|c|c}
            Input & $\xb$ & $\R^\nx$\\\hline
            Output & $\hat{\y}$ & $\R^\ny$\\\hline
            (Hidden) State & $\z$ & $\R^\nz$\\\hline
            Parameters & $\theta(s)$ & $\R^\nt$\\\hline
            Neural Vector Field & $f_{\theta(s)}$ & $\R^\nz$ 
            \\\hline
            Input Network & $h_x$ & $\R^\nx \ra \R^\nz$\\\hline
            Output Network & $h_y$ & $\R^\nz \ra \R^\ny$\\
        \end{tabular}
    \end{minipage}
\end{CatchyBox}
\paragraph{\fcircle[fill=wine]{3pt} Depth--variance}
Vanilla Neural ODEs \citep{chen2018neural} cannot be considered the deep limit of ResNets. We discuss the subtleties involved, uncovering a formal optimization problem in functional space as the price to pay for true depth--variance. Obtaining its solution leads to two novel variants of Neural ODEs: a Gal$\ddot{\text{e}}$rkin--inspired spectral discretization (GalNODE) and a piecewise--constant model. GalNODEs are showcased on a task involving a loss distributed on the depth--domain, requiring the introduction of a \textit{generalized} version of the adjoint in \citep{chen2018neural}.
\paragraph{\fcircle[fill=deblue]{3pt} Augmentation strategies}
The augmentation idea of ANODEs \citep{dupont2019augmented} is taken further and generalized to novel dynamical system--inspired and parameter efficient alternatives, relying on different choices of $h_x$ in (\ref{eq:1}). These approaches, which include \textit{input--layer} and \textit{higher--order} augmentation, are verified to be more effective than existing methods in terms of performance and parameter efficiency.
\paragraph{\fcircle[fill=olive]{3pt} Beyond augmentation: data--control and adaptive--depth}
We unveil that although important, augmentation is not always necessary in challenging tasks such as learning \textit{reflections} or \textit{concentric annuli} \citep{dupont2019augmented}. To start, we demonstrate that depth--varying vector fields alone are sufficient in dimensions greater than one. We then provide theoretical and empirical results motivating two novel Neural ODE paradigms: \textit{adaptive--depth}, where the integration bound is itself determined by an auxiliary neural network, and \textit{data--controlled}, where $f_{\theta(s)}$ is conditioned by the input data $\xb$, allowing the ODE to learn a \textit{family} of vector fields instead of a single one. Finally, we warn against input networks $h_x$ of the multilayer, nonlinear type, as these can make Neural ODE flows \textit{superfluous}.
 

%% file: 2_Background.tex
\vspace{-3mm}
\section{Continuous--Depth Models}
\vspace{-3mm}
\paragraph{A general formulation}
In the context of Neural ODEs we suppose to be given a stream of input--output data $\{(\xb_k,\y_k)\}_{k\in\cK}$ (where $\cK$ is a linearly--ordered finite subset of $\N$). The inference of Neural ODEs is carried out by solving the \textit{inital value problem} (IVP) \eqref{eq:1}, i.e.
\setlength{\abovedisplayskip}{1pt}
\setlength{\belowdisplayskip}{1pt}
\begin{equation*}
    \hat{\y}(S) = h_y\left(h_x(\xb) + \int_\cS f_{\theta(\tau)}(\tau,\xb,\z(\tau))\dd \tau\right)
\end{equation*}
 Our degree of freedom, other than $h_x$ and $h_y$, in the Neural ODE model is the choice of the parameter $\theta$ inside a given pre-specified class $\cW$ of functions $\cS\to\R^\nt$.
\vspace{-4mm}
\paragraph{Well--posedness}
If $f_{\theta(s)}$ is Lipschitz, for each $\xb_k$ the initial value problem in \eqref{eq:1} admits a unique solution $\z$ defined in the whole $\cS$. If this is the case, there is a mapping $\phi$ from $\cW\times\R^{\nx}$ to the space of absolutely continuous functions $\cS\mapsto\R^{\nz}$ such that $\z_k:= \phi(\xb_k, \theta)$ satisfies the ODE in \eqref{eq:1}. This in turn implies that, for all $k\in\cK$, the map
$(\theta,\xb_k, s)\mapsto \gamma( s,\xb_k,\theta) := h_y\big(\phi(\theta ,\xb_k)(s)\big)$ 
satisfies $\hat {\y} = \gamma(\theta,\xb_k,s)$.
For compactness, for any $s\in\cS$, we denote $\phi(\theta,\xb_k)(s)$ by $\phi_s(\theta,\xb_k)$.
\paragraph{Training: optimal control}
\citep{chen2018neural} treated the training of constant--parameters Neural ODE (i.e. $\cW$ is the space of constant functions) considering only \textit{terminal} loss functions depending on the terminal state $\z(S)$. However, in the framework of Neural ODEs, the latent state evolves through a continuum of layers steering the model output $\hat \y(s)$ towards the label. It thus makes sense to introduce a loss function also distributed on the whole depth domain $\cS$, e.g.
\begin{equation}\label{eq:2}
    \ell := L(\z(S)) + \int_{\cS}l(\tau, \z(\tau))\dd \tau
\end{equation}
The training can be then cast into the \textit{optimal control} \citep{pontryagin1962mathematical} problem
\begin{equation}\label{eq:3}
    \begin{aligned}
        \min_{\theta\in\cW}  &\frac{1}{|\cK|}\sum_{k\in\cK}\ell_k \\
        \text{subject to}~~
        &\dot{\z}(s) =  f_{\theta (s)}\left(s, \xb_k, \z(s) \right)~~s \in \cS\\
        & \z(0) = h_x(\xb_k),~~\hat{\y}(s) = h_y(\z(s))
    \end{aligned},~~\forall k\in\cK
\end{equation}
solved by gradient descent. Here, if $\theta$ is constant, the gradients can be computed with $\cO(1)$ memory efficiency by generalizing the adjoint sensitivity method in \citep{chen2018neural}.
\begin{restatable}[Generalized Adjoint Method]{proposition}{GenAdj}\label{thm:GenAdj}
Consider the loss function \eqref{eq:2}. Then,
    \[
        \frac{\dd\ell}{\dd\theta} = \int_\cS \ab^\top(\tau)\frac{\partial f_{\theta}}{\partial \theta}\dd\tau~~\text{where $\ab(s)$ satisfies}
        ~~\left\{
        \begin{matrix*}[l]
            \dot{\ab}^\top(s) = -\ab^\top(s) \frac{\partial f_{\theta}}{\partial \z} - \frac{\partial l}{\partial \z}\\
            \ab^\top(S) =  \frac{\partial L}{\partial \z(S)}
        \end{matrix*}\right.
    \]
\end{restatable}
Supplementary material contains additional insights on the choice of activation, training regularizers and approximation capabilities of Neural ODEs, along with a detailed derivation of the above result.%
%

%% file: 3_NeuralDEs.tex
\section{Depth-Variance: Infinite Dimensions for Infinite Layers}\label{sec:dissecting}
\paragraph{Bring residual networks to the deep limit} \textit{Vanilla} Neural ODEs, as they appear in the original paper \citep{chen2018neural}, cannot be fully considered the \textit{deep limit} of ResNets. In fact, while each residual block is characterized by its own parameters vector $\theta_s$, the authors consider model $\dot{\z} = f_\theta(s, \z(s))$ where the depth variable $s$ enters in the dynamics \textit{per se}\footnote{In practice, $s$ is often concatenated to $\z$ and fed to $f_\theta$.} rather than in the map $s\mapsto\theta(s)$. The first attempt to pursue the true deep limit of ResNets is the \textit{hypernetwork} approach of \citep{zhang2019anodev2} where another neural network parametrizes the dynamics of $\theta (s)$.

However, this approach is not backed by any theoretical argument and it exhibits a considerable parameter inefficiency, as it generally scales polynomially in $\nt$. We adopt a different approach, setting out to tackle the problem theoretically in the general formulation. Here, we uncover an optimization problem in functional space, solved by a direct application of the adjoint sensitivity method in infinite-dimensions. We then introduce two parameter efficient depth--variant Neural ODE architectures based on the solution of such problem: \textit{Gal$\ddot{e}$rkin} Neural ODEs and \textit{Stacked} Neural ODEs.
\paragraph{Gradient descent in functional space \fcircle[fill=wine]{2pt}} %
When the model parameters are depth--varying, $\theta:\cS\rightarrow\R^\nt$, the nonlinear optimization problem \eqref{eq:3} should be in principle solved by iterating a gradient descent algorithm in a functional space \citep{smyrlis2004local}, e.g. $\theta_{k+1}(s) = \theta_k(s) - \eta{\delta\ell_k}/{\delta\theta}(s)$ once the Gateaux derivative ${\delta\ell_k}/{\delta\theta}(s)$ is computed. Let $\bL_2(\cS\rightarrow\R^\nt)$ be the space of square--integrable functions $\cS\rightarrow\R^\nt$.  Hereafter, we show that if $\theta(s)\in\cW:=\bL_2(\cS\rightarrow\R^\nt)$, then the loss sensitivity to $\theta(s)$ can be computed through the adjoint method. 
\begin{restatable}[Infinite--Dimensional Gradients]{theorem}{InfAdj}\label{thm:2}
    Consider the loss function \eqref{eq:2} and let $\theta(s)\in\mathbb{L}_2(\cS\rightarrow\R^\nt)$. Then, sensitivity of $\ell$ with respect to $\theta(s)$ (i.e. directional derivative in functional space) is
    $$
        \frac{\delta\ell}{\delta \theta (s)} = \ab^\top(s)\frac{\partial f_{\theta(s)}}{\partial \theta(s)}~~\text{where $\ab(s)$ satisfies}
        ~~\left\{
        \begin{matrix*}[l]\small
            \dot{\ab}^\top(s) = -\ab^\top(s)\frac{\partial f_{\theta(s)}}{\partial \z} - \frac{\partial l}{\partial \z}\\
            \ab^\top(S) =  \frac{\partial L}{\partial \z(S)}
        \end{matrix*}\right.
    $$
\end{restatable}
Note that, although Theorem \ref{thm:2} provides a constructive method to compute the loss gradient in the \textit{infinite--dimensional} setting, its implementation requires choosing a finite dimensional approximation of the solution. We offer two alternatives: a \textit{spectral discretization} approach relying on reformulating the problem on some functional bases and a \textit{depth discretization} approach.
\paragraph{Spectral discretization: \textit{Gal\"erkin Neural ODEs} \fcircle[fill=wine]{2pt}} The idea is to expand $\theta(s)$ on a complete orthogonal basis of a predetermined subspace of $\mathbb{L}_2(\cS\rightarrow\R^\nt)$ and truncate the series to the $m$-th term:
    \[
        \theta(s) = \sum_{j = 1}^m\alpha_j\odot\psi_j(s)
    \]
In this way, the problem is turned into finite dimension and the training will aim to optimize the parameters $\alpha=(\alpha_1,\dots,\alpha_m)\in\R^{m\nt}$ whose gradients can be computed as follows
\begin{restatable}[Spectral Gradients]{corollary}{Cone}\label{thm:cor1}
    Under the assumptions of Theorem \ref{thm:2}, if $\theta(s) = \sum_{j=1}^m \alpha_j\odot\psi_j(s)$,%
    $$
        \frac{\dd \ell}{\dd \alpha}=\int_\cS \ab^\top(\tau)\frac{\partial f_{\theta(s)}}{\partial\theta (s)}\psi(\tau)\dd \tau,~~~~\psi = (\psi_1,\dots,\psi_m)
    $$
\end{restatable}
\paragraph{Depth discretization: \textit{Stacked Neural ODEs} \fcircle[fill=wine]{2pt}} An alternative approach to parametrize $\theta(s)$ is to assume it piecewise constant in $\cS$, i.e. $\theta(s) = \theta_i~~\forall s \in [s_i,s_{i+1}]$ and $\cS = \bigcup_{i=0}^{p-1} [s_i,s_{i+1}]$. 
It is easy to see how evaluating this model is equivalent to \textit{stacking} $p$ Neural ODEs with constant parameters,
$$
    \z(S) = h_x(\xb) + \sum_{i=0}^{p-1}\int_{s_1}^{s_{i+1}}f_{\theta_i}(\tau, \xb, \z(\tau))\dd \tau
$$
Here, the training is carried out optimizing the resulting $p\nt$ parameters using the following:
\begin{restatable}[Stacked Gradients]{corollary}{Ctwo}\label{thm:cor2}
    Under the assumptions of Theorem \ref{thm:2}, if $\theta(s) =\theta_i~\forall s \in [s_i,s_{i+1}]$,%
    $$
        \frac{\dd \ell}{\dd \theta_i}=-\int_{s_{i+1}}^{s_{i}} \ab^\top(\tau)\frac{\partial f_{\theta_i}}{\partial\theta_i}\dd \tau~~\text{where $\ab(s)$ satisfies}
        ~~\left\{
        \begin{matrix*}[l]
            \dot{\ab}^\top(s) = -\ab^\top(s)\frac{\partial f_{\theta_i}}{\partial \z} - \frac{\partial l}{\partial \z}~~~s\in[s_i,s_{i+1}]\\
            \ab^\top(S) =  \frac{\partial L}{\partial \z(S)}
        \end{matrix*}\right.
    $$
\end{restatable}

The two approaches offer different perspectives on the problem of parametrizing the evolution of $\theta(s)$; while the spectral method imposes a stronger prior to the model class, based on the chosen bases (e.g. Fourier series, Chebyshev polynomials, etc.) the depth--discretization method allows for more freedom. {Further details on proofs, derivation and implementation of the two models are given in the Appendix.}
\paragraph{Tracking signals via depth--variance}
\begin{wrapfigure}[14]{l}{0.5\textwidth}
    \vspace{-2mm}
    \includegraphics[scale=.95]{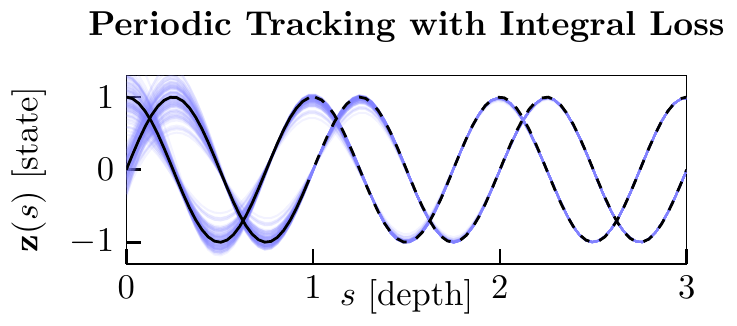}
    \vspace{-7mm}
    \caption{Gal\"erkin Neural ODEs trained with integral losses accurately recover periodic signals. Blue curves correspond to different initial conditions and converge asymptotically to the reference desired trajectory.}
    \label{fig:1}
\end{wrapfigure}
Consider the problem of tracking a periodic signal $\beta(s)$. We show how this can be achieved without introducing additional inductive biases such as \citep{greydanus2019hamiltonian} through a synergistic combination of a two--layer Gal\"erkin Neural ODEs and the generalized adjoint with integral loss $l(s):=\|\beta(s)-\z(s)\|^2_2$. The models, trained in $s\in[0,1]$ generalize accurately in extrapolation, recovering the dynamics. 
\begin{figure}[b]
    \centering
    \includegraphics[scale=.95]{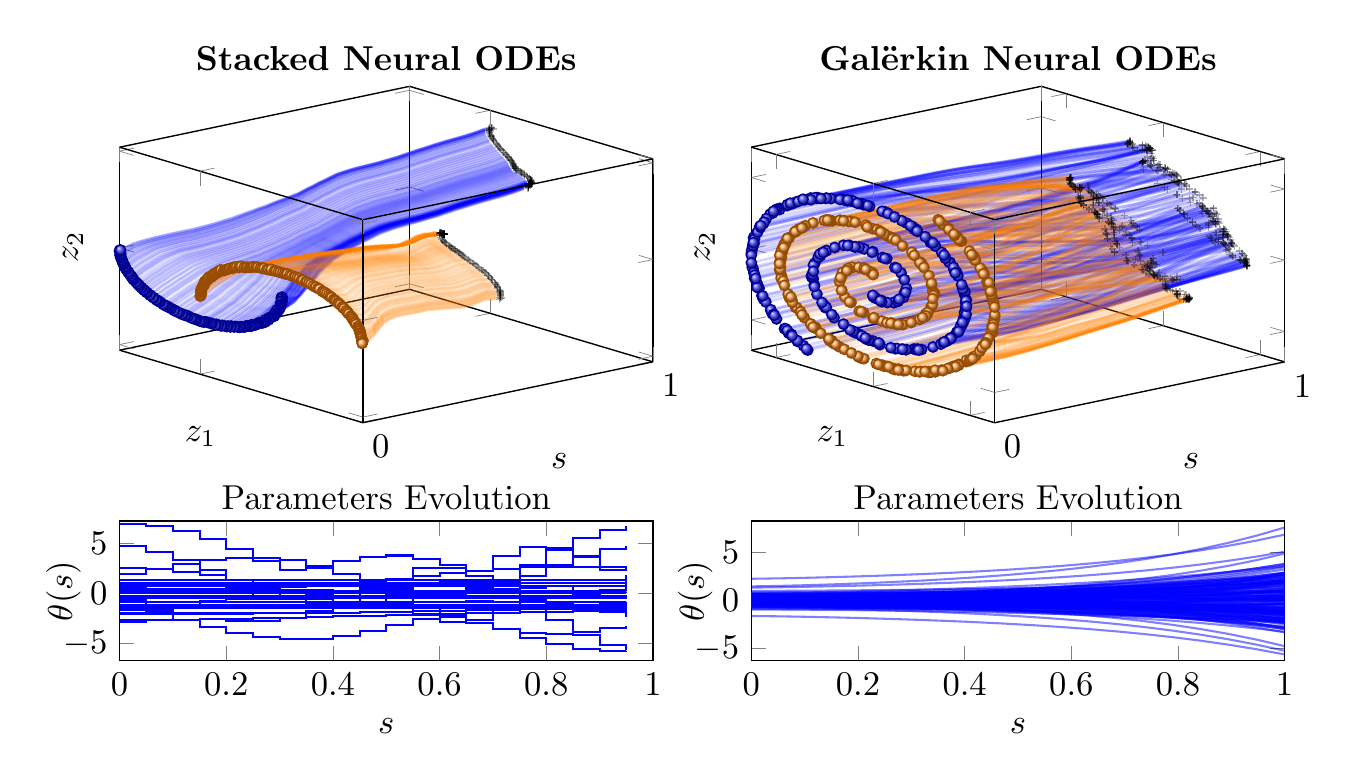}
    \vspace{-8mm}
    \caption{Gal\"erkin and Stacked parameter-varying Neural ODE variants. Depth flows (Above) and evolution of the parameters (Below).}
    \label{fig:2}
\end{figure}
Fig.\ref{fig:2} showcases the depth--dynamics of $\theta(s)$ for Gal\"erkin and Stacked variants trained to solve a simple binary classification problem. Additional insights and details are reported in Appendix.

\begin{tcolorbox}\centering
    Depth--variance brings Neural ODEs closer to the ideal continuum of neural network layers with untied weights, enhancing their expressivity. 
\end{tcolorbox}

%% file: 4_Augmentation.tex
\vspace{-3mm}
\section{Augmenting Neural ODEs}\label{sec:augmenting}
\textit{Augmented Neural ODEs} (ANODEs) \citep{dupont2019augmented} propose solving the \textit{initial value problem} (IVP) in a higher dimensional space to limit the complexity of learned flows, i.e. having $\nz > \nx$. 
The proposed approach of the seminal paper relies on initializing to zero the $n_a := \nz-\nx$ augmented dimensions: $\z(0) = [\xb,\mathbb{0}]$.
We will henceforth refer to this augmentation strategy as \textit{$0$--augmentation}. In this section we discuss alternative augmentation strategies for Neural ODEs that match or improve on $0$--augmentation in terms of performance or parameter efficiency.
\paragraph{Input--layer augmentation \fcircle[fill=deblue]{2pt}}
Following the standard deep learning approach of increasing layer width to achieve improved model capacity, $0$--augmentation can be generalized by introducing an input network $h_x:\R^\nx \rightarrow \R^\nz$ to compute $\z(0)$:
    \begin{equation}\label{eq:IL_aug}
            \z(0) = h_x(\xb)
    \end{equation}
leading to the general formulation of \eqref{eq:1}. This approach (\ref{eq:IL_aug}) gives the model more freedom in determining the initial condition for the IVP instead of constraining it to a concatenation of $\xb$ and $\mathbb{0}$, at a small parameter cost if $h_x$ is, e.g., a linear layer. We refer to this type of augmentation as \textit{input layer} (IL) augmentation and to the model as \textit{IL--Neural ODE} (IL--NODE). 

Note that 0-augmentation is compatible with the general IL formulation, as it corresponds to
$$
    \xb\mapsto (\xb, \mathbb{0}):=h_x(\xb)
$$
In applications where maintaining the structure of the first $\nx$ dimensions is important, e.g. approximation of dynamical systems, a parameter efficient alternative of (\ref{eq:IL_aug}) can be obtained by modifying the input network $h_x$ to only affect the additional $n_a$ dimensions, i.e. $h_x := [\xb,\xi(\xb)]$, $\xi:\R^\nx\rightarrow\R^{n_a}$.
%
\paragraph{Higher--order Neural ODEs \fcircle[fill=deblue]{2pt}}
Further parameter efficiency can be achieved by lifting the Neural ODEs to higher orders. For example, let $\z(s) = [\z_q(s), \z_p(s)]$ a second--order Neural ODE of the form:
\begin{equation}\label{eq:2ord}
    \ddot{\z}_q(s) = f_{\theta(s)}(s, \z(s)).
\end{equation}
equivalent to the first--order system 
\begin{equation}\label{eq:HO_aug}
    \begin{aligned}
        \dot{\z}_q(s) &= \z_p(s)\\
        \dot{\z}_p(s) &= f_{\theta(s)}(s, \z_q(s),\z_p(s))
    \end{aligned}
\end{equation}
The above can be extended to higher--order Neural ODEs as
\begin{equation}\label{eq:ho}
    \frac{\dd^n \z^1}{\dd s^n} = f_{\theta(s)}\left(s,\z,\frac{\dd \z^1}{\dd s},\cdots,\frac{\dd^{n-1} \z^1}{\dd s^{n-1}}\right),~~~~\z=[\z^1,\z^2,\dots,\z^{n}], ~\z^i\in\R^{\nz/n}
\end{equation}
or, equivalently, $\dot \z^i(s) = \z^{i+1}(s)$, $\dot\z^n(s) = f_{\theta(s)}(s, \z(s))$. Note that the parameter efficiency of this method arises from the fact that $f_{\theta(s)}:\R^{\nz}\rightarrow\R^{\nz/n}$ instead of $\R^{\nz}\rightarrow\R^{\nz}$. A limitation of system (\ref{eq:HO_aug}) is that a naive extension to second--order requires a number of augmented dimensions $n_a = \nx$. To allow for flexible augmentations of few dimensions $n_a<\nx$, the formulation of second--order Neural ODEs can be modified to \textit{select} only a few dimensions to have higher order dynamics. We include formulation and additional details of \textit{selective} higher--order augmentation in the supplementary material. Finally, higher--order augmentation can itself be compatible with input--layer augmentation.
\begin{table*}[b]\label{tab:mnist}
    \centering
    \setlength\tabcolsep{4pt}
        \begin{tabular}{l|c|c|c|c|}
            \toprule
            & NODE & ANODE & \color{blue!70!white} IL-NODE & \color{blue!70!white}2nd--Ord. \\
            \midrule
            &\tiny{MNIST \big | CIFAR}~~&\tiny{MNIST \big | CIFAR}~~&\tiny{MNIST \big | CIFAR}~~&\tiny{MNIST \big | CIFAR}~~\\\hline
            Test Acc. &  $96.8$ \big | $58.9$& $98.9$ \big | $70.8$&  ~{\color{blue!70!white}$99.1$} \big | {\color{blue!70!white}$\mathbf{73.4}$}&  {\color{blue!70!white}$\mathbf{99.2}$} \big | {\color{blue!70!white}$72.8$}\\
            NFE & $98$ \big | $93$& ~~$71$ \big | $169$&\color{blue!70!white} $44$ {\color{black}\big |} $65$&~\color{blue!70!white} $\mathbf{43}$ {\color{black}\big |} $\mathbf{59}$\\
            Param.[K] & $21.4$ {\color{black}\big |} $37.1$ &$20.4$ {\color{black}\big |} $35.0$ &\color{blue!70!white}$20.7$ {\color{black}\big |} $36.1$ &~\color{blue!70!white}$\mathbf{20.0}$ {\color{black}\big |} $\mathbf{34.6}$\\
            \bottomrule
        \end{tabular}
    
    \caption{\footnotesize Mean test results across 10 runs on MNIST and CIFAR. We report the mean NFE at convergence. Input layer and higher order augmentation improve task performance and preserve low NFEs at convergence.}
    \label{tab:allresone}
\end{table*}
\paragraph{Revisiting results for augmented Neural ODEs}
In higher dimensional state spaces, such as those of image classification settings, the benefits of augmentation become subtle and manifest as performance improvements and a lower \textit{number of function evaluations} (NFEs) \citep{chen2018neural}. We revisit the image classification experiments of \citep{dupont2019augmented} and evaluate four classes of depth--invariant Neural ODEs: namely, vanilla (no augmentation), ANODE (0--augmentation), IL-NODE (input--layer augmentation), and second--order. The input network $h_x$ is composed of a single, linear layer. Main objective of these experiments is to rank the efficieny of different augmentation strategies; for this reason, the setup does not involve hybrid or composite Neural ODE architectures and data augmentation.

The results for five experiments are reported in Table \ref{tab:mnist}. IL--NODEs consistently preserve lower NFEs than other variants, whereas second--order Neural ODEs offer a parameter efficient alternative. The performance gap widens on CIFAR10, where the disadvantage of fixed $0$ initial conditions forces $0$--augmented Neural ODEs into performing a high number of function evaluations.

It should be noted that prepending an input multi--layer neural network to the Neural ODE was the approach chosen in the experimental evaluations of the original Neural ODE paper \citep{chen2018neural} and that \citep{dupont2019augmented} opted for a comparison between no input layer and $0$--augmentation. However, a significant difference exists between architectures depending on the depth and expressivity of $h_x$. Indeed, utilizing non--linear and multi--layer input networks can be detrimental, as discussed in Sec. \ref{sec:no_aug}. 
\begin{tcolorbox}\centering
    Augmentation relieves Neural ODEs of their expressivity limitations. Learning initial conditions improves on 0--augmentation in terms of performance and NFEs.
\end{tcolorbox}

%% file: 5_Non_Augmentation.tex
\section{Beyond Augmentation: Data--Control and Depth--Adaptation}\label{sec:no_aug}
Augmentation strategies are not always necessary for Neural ODEs to solve challenging tasks such as \textit{concentric annuli} \citep{dupont2019augmented}. While it is indeed true that two distinct trajectories can never intersect in the state--space in the one--dimensional case, this does not necessarily hold in general. In fact, dynamics in the first two spatial dimensions are substantially different e.g no chaotic behaviors are possible \citep{khalil2002nonlinear}. In the two--dimensions of $\R^2$ (and so in $\R^n$), infinitely wider than $\R$, distinct trajectories of a time--varying process can well intersect in the state--space, provided that they do not pass through the same point at the same time \citep{khalil2002nonlinear}. This implies that, in turn, depth--varying models such as Gal$\ddot{\text{e}}$rkin Neural ODEs can solve these tasks in all dimensions but $\R$. 

Starting from the one--dimensional case, we propose new classes of models allowing Neural ODEs to perform challenging tasks such as approximating \textit{reflections} \citep{dupont2019augmented} without the need of any augmentation. 
\begin{figure}[t]
    \centering
    \input{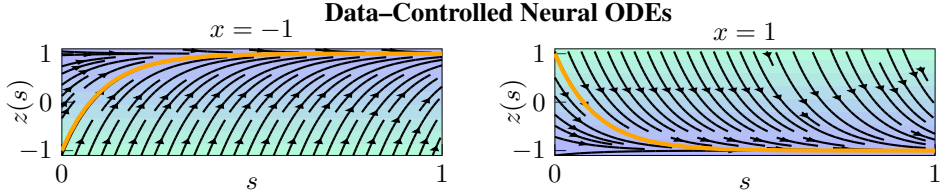}
    \vspace{-2mm}
    \caption{Depth trajectories over vector field of the \textit{data--controlled} neural ODEs (\ref{eq:lin_sys}) for $x=1,~x=-1$. The model learns a family of vector fields conditioned by the input $x$ to approximate $\varphi(x)$.}
    \label{fig:lti}
\end{figure}
%
\subsection{Data--controlled Neural ODEs \fcircle[fill=sand]{2pt}}          
We hereby derive a new class of models, namely \textit{data--controlled Neural ODEs}.

To introduce the proposed approach, we start with an analytical result regarding the approximation of reflection maps such as $\varphi(x) = -x$. The proof provides a design recipe for a simple handcrafted ODE capable of approximating $\varphi$ with arbitrary accuracy by leveraging input data $x$. We denote the conditioning of the vector field with $x$ necessary to achieve the desired result as \textit{data--control}. 

This result highlights that, through data--control, Neural ODEs can arbitrarily approximate $\varphi$ without augmentation, providing a novel perspective on existing results about expressivity limitations of continuous models \citep{dupont2019augmented}. The result is the following:
\begin{restatable}{proposition}{san}\label{th:san}
    For all $\epsilon>0$, $x\in\R$ there exists a parameter $\theta>0$ such that 
    \begin{equation}
        \left|\varphi(x) - z(1)\right|<\epsilon,
    \end{equation}
    where $z(1)$ is the solution of the Neural ODE
    \begin{equation}\label{eq:lin_sys}
        \left\{
            \begin{matrix*}[l]
                \dot{z}(s) = -\theta(z(s)+x)\\
                z(0) = x
            \end{matrix*},~~s\in[0,1]
        \right.~.
    \end{equation}
\end{restatable}
The proof is reported in the Appendix. Fig. (\ref{fig:lti}) shows a version of model \eqref{eq:lin_sys} where $\theta$ is trained with standard backpropagation. This model is indeed able to closely approximate $\varphi(x)$ without augmentation, confirming the theoretical result.
From this preliminary example, we then define the general \textit{data--controlled} Neural ODE as
\begin{equation}\label{eq:cnode}
    \begin{aligned}
        \dot{\z}(s) &= f_{\theta(s)}(s, {\color{olive}\xb}, \z(s))\\
        \z(0) &= h_x({\color{olive}\xb})
    \end{aligned}.
\end{equation}
Model \eqref{eq:cnode} incorporates input data $\xb$ into the vector field, effectively allowing the ODE to learn a \textit{family} of vector fields instead of a single one. Direct dependence on $\xb$ further constrains the ODE to be smooth with respect to the initial condition, acting as a regularizer. Indeed, in the experimental evaluation at the end of Sec. \ref{sec:no_aug}, data--controlled models recover an accurate decision boundary. Further experimental results with the latter general model on the representation of $\varphi$ are reported in the Appendix.

It should be noted that \eqref{eq:cnode} does not require explicit dependence of the vector field on $\xb$. Computationally, $\xb$ can be passed to $f_{\theta(s)}$ in different ways, such as through an additional embedding step. In this setting, data--control offers a natural extension to conditional Neural ODEs.
\vspace{-2mm}
\paragraph{Data--control in normalizing flows}
\begin{wrapfigure}[25]{l}{0.5\textwidth}\label{2NFE}
    \vspace{-5mm}
    \centering
    \includegraphics[width=0.5\textwidth]{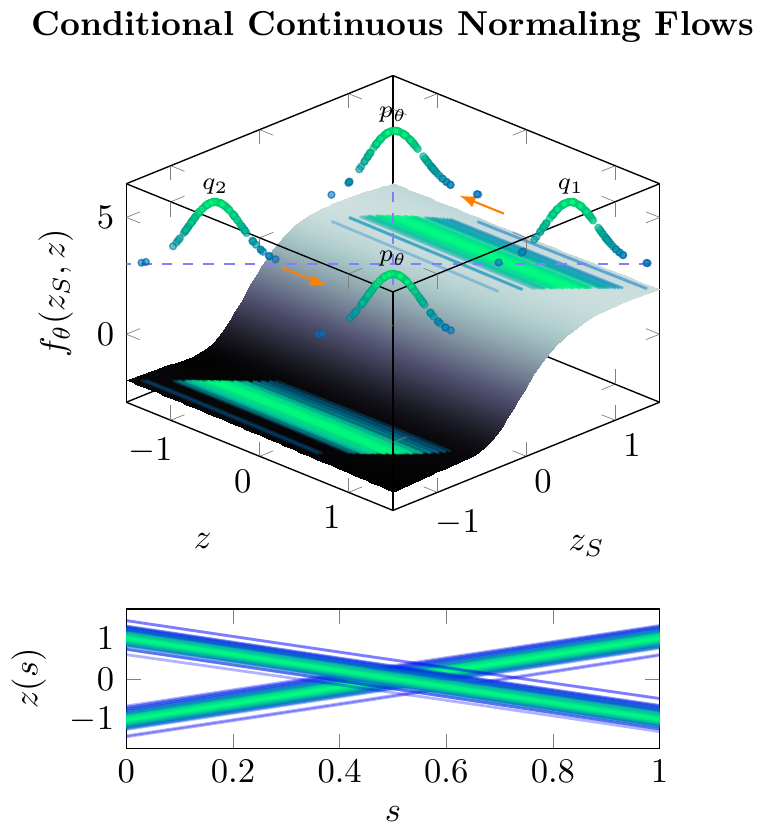}
    \vspace{-8mm}
    \caption{Data--controlled CNFs can morph prior distributions into distinct posteriors to produce conditional samples. This task often requires crossing trajectories and is not possible with vanilla CNFs.}
    \label{fig:dccnf}
\end{wrapfigure}
Conditional variants of generative models can be guided to produce samples of different characteristics depending on specific requirements. Data--control can be leveraged to obtain a conditional variant of continuous normalizing flows (CNFs) \citep{chen2018neural}. Here, we consider the standard setting of learning an unknown data distribution $p(\xb)$ given samples $\{\xb_k\}_{k\in\cK}$ through a parametrized function $p_\theta$.
\textit{Continuous normaling flows} (CNFs) \citep{chen2018neural, grathwohl2018ffjord, finlay2020train} obtain $p_\theta$ by change of variables using the flow of an ODE to warp a (known) prior distribution $q(\z)$, i.e. $\log p_\theta(\xb) = \log q(\phi_S(\xb)) + \log\det|\nabla\phi_S(\xb)|$ where the $\log$ determinant of the Jacobian is computed via the fluid mechanics identity $$\frac{\dd}{\dd s}\log\det|\nabla\phi_s(\xb)| = \nabla\cdot f_{\theta(t)}(s, \phi_s(\xb)),$$
\citep{villani2003topics}. CNFs are trained via maximum--likelihood, i.e by minimizing the Kullback--Leibler divergence between $p$ and $p_\theta$, or equivalently $\ell:=-1/|\cK|\sum_k \log p_\theta(\xb_k)$. A CNF can be then used as generative model for $p_\theta(\xb)$ by sampling the known distribution $\z_S\sim q(\z_S)$ and evolve $\z_S$ backward in the depth domain:
\[
    \z(0) = \z_S + \int_S^0 f_\theta(s)(s, \z(s))\dd s
\]
In this context, introducing data--control into $f_{\theta}$ allows the CNF to be conditioned with data or task information. Data--controlled CNFs can thus be used in multi--objective generative tasks e.g using a single model to sample from $N$ different distribution $p_\theta$ by warping $N$ predetermined known distributions $q_i$. 
We train one--dimensional, data--controlled CNFs to approximate two different data distributions $p_1,~p_2$ by sampling from two distinct priors $q_1,~q_2$ and conditioning the vector field with the samples $z_S$ of the prior distributions, i.e.
\[
    \dot z(s) = f_\theta(z_S, z(s)),~ ~~z_S\sim q_1\text{ or }z_S\sim q_2
\]
Fig \ref{fig:dccnf} shows how data--controlled CNFs are capable of conditionally sampling from two normal target data distributions. In this case we selected $p_1, p_2$ as univariate normal distributions with mean $-1$ and $1$, respectively and $q_1\equiv p_2,~q_2\equiv p_1$. The resulting learned vector field strongly depends on the value of the prior sample $z_S$ and it is almost constant in $z$, meaning that the prior distributions are just shifted almost rigidly along the flow in a direction determined by the initial condition. 
This task is inaccessible to standard CNFs as it requires crossing flows in $z$. Indeed, the proposed benchmark represents a density estimation analogue to the crossing trajectories problem. %
\vfill
\subsection{Adaptive--Depth Neural ODEs \fcircle[fill=sand]{2pt}}
\begin{wrapfigure}[12]{r}{0.5\textwidth}
    \vspace{-4mm}
    \centering
    \input{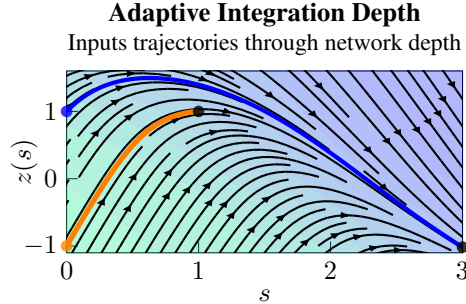}
    \vspace{-4mm}
    \caption{Depth trajectories over vector field of the
\textit{adaptive–-depth} Neural ODEs. The reflection map
can be learned by the proposed model. The key is
to assign different integration times to the inputs,
thus not requiring the intersection of trajectories.}
    \label{fig:adapt_time}
\end{wrapfigure}

Let us come back to the approximation of $\varphi(x)$. Indeed, without incorporating input data into $f_{\theta(s)}$, it is not possible to realize a mapping $x\mapsto \phi_s(x)$ mimicking $\varphi$ due to the topology preserving property of the flows. Nevertheless, a Neural ODE can be employed to approximate $\varphi(x)$ without the need of any \textit{crossing trajectory}. In fact, if each input is integrated for in a different depth domain, $\cS(x) = [0, s^*_x]$, it is possible to learn $\varphi$ without crossing flows as shown in Fig. \ref{fig:adapt_time}. 
In general, we can use a hypernetwork $g$ trained to learn the integration depth of each sample. In this setting, we define the general \textit{adaptive depth} class as Neural ODEs performing the mapping $\xb\mapsto\phi_{g_\omega (\xb)}(\xb)$, i.e. leading to
\begin{equation*}
    \hat{\y} = h_y\left( h_x(\xb) + \int_0^{g_{\omega}(\xb)}f_{\theta(s)}(\tau, \xb, \z(\tau))\dd\tau\right),
\end{equation*}
where $g_\omega:\R^\nx\times\R^\nw\rightarrow\R$ is a neural network with trainable parameters $\omega$. Supplementary material contains details on differentiation under the integral sign, required to back--propagate the loss gradients into $\omega$.
\subsection{Additional Results}
\paragraph{Experiments of non--augmented models}
We inspect the performance of different Neural ODE variants: depth--invariant, depth--variant with $s$ concatenated to $\z$ and passed to the vector field, Gal$\ddot{\text{e}}$rkin Neural ODEs and data--controlled. The concentric annuli \citep{dupont2019augmented} dataset is utilized, and the models are qualitatively evaluated based on the complexity of the learned flows and on how accurately they extrapolate to unseen points, i.e. the learned decision boundaries. For Gal$\ddot{\text{e}}$rkin Neural ODEs, we choose a Fourier series with $m = 5$ harmonics as the eigenfunctions $\psi_k$, $k = 1, \dots, 5$ to compute the parameters $\theta(s)$, as described in Sec. \ref{sec:dissecting}.
\begin{tcolorbox}\centering
\small
    Data--control allows Neural ODEs to learn a family of vector fields, conditioning on input data information. Depth--adaptation sidesteps known expressivity limitations of continuous--depth models.
\end{tcolorbox}
\paragraph{Mind your input networks}
\begin{wrapfigure}[13]{l}{0.5\textwidth}
    \centering
    \vspace{-6mm}
    \includegraphics[width=.95\linewidth]{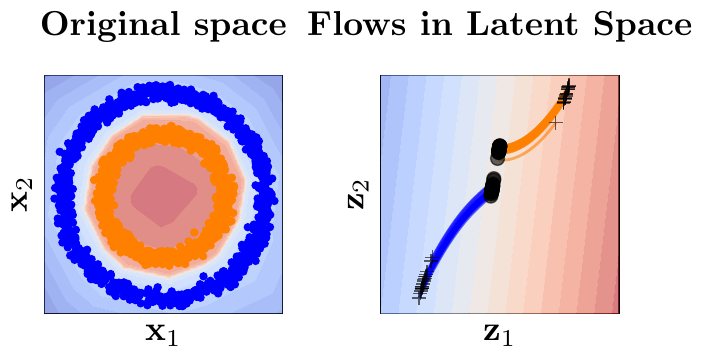}
    \vspace{-3mm}
    \caption{Solving concentric annuli without augmentation by prepending a nonlinear transformation performed by a two--layer fully--connected network.}
    \label{fig:il_noaug}
\end{wrapfigure}
An alternative approach to learning maps that prove to be challenging to approximate for vanilla Neural ODEs involves solving the ODE in a latent state space. Fig. \ref{fig:il_noaug} shows that with no augmentation, a network composed by a two fully--connected layers with non--linear activation followed by a Neural ODE can solve the concentric annuli problem. However, the flows learned by the Neural ODEs are superfluous: indeed, the clusters were already linearly separable after the first non--linear transformation. This example warns against superficial evaluations of Neural ODE architectures preceded or followed by several layers of non--linear input and output transformations. In these scenarios, the learned flows risk performing unnecessary transformations and in pathological cases can collapse into a simple identity map. To sidestep these issues, we propose visually inspecting trajectories or performing an ablation experiment on the Neural ODE block.
\newpage
\begin{figure*}[!t]
    \centering
    \includegraphics{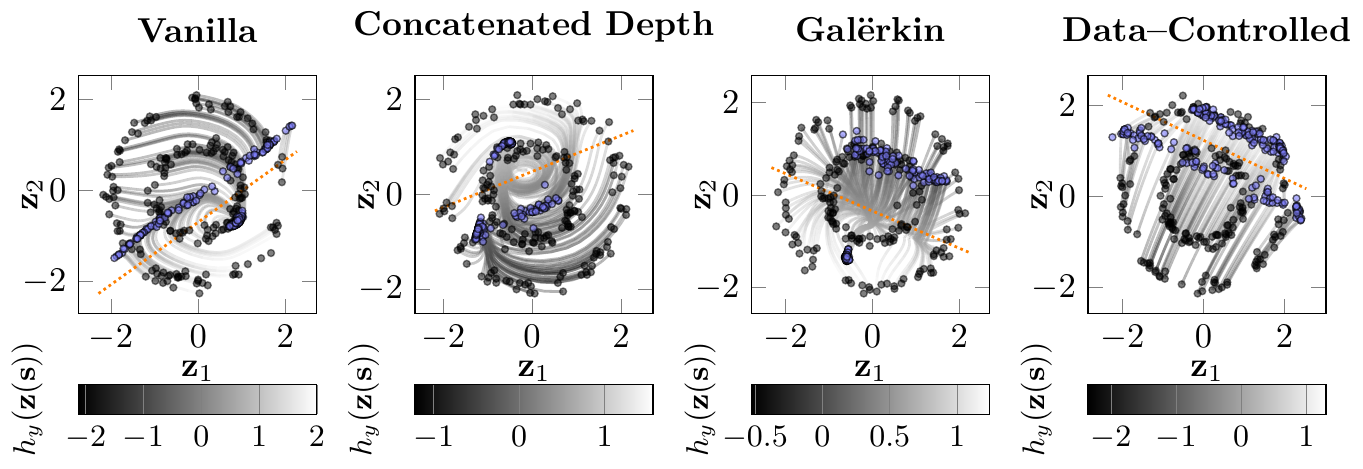}
    \par
    \vspace{-2mm}
    \caption{Depth-flows of the data in the state--space. The resulting decision boundaries of output linear layer $h_y$ are indicated by the dotted orange line.}
    \label{fig:dec_bound}
\end{figure*}
%
%
%

%% file: 6_Related_Work.tex
\vspace*{-10mm}
\section{Related Work}
We include a brief history of classical approaches to dynamical system--inspired deep learning.
\paragraph{A brief historical note on continuous deep learning}
Continuous neural networks have a long history that goes back to continuous time variants of recurrent networks \citep{cohen1983absolute}. Since then, several works explored the connection between dynamical systems, control theory and machine learning \citep{zhang2014comprehensive,li2017maximum,lu2017expressive,weinan2017proposal}. \citep{marcus1989stability} provides stability analyses and introduces delays. Many of these concepts have yet to resurface in the context of Neural ODEs. \cite{haber2017stable} analyzes ResNet dynamics and links stability with robustness. Injecting stability into neural networks has inspired the design of a series of architectures \citep{chang2019antisymmetricrnn,haber2019imexnet,NIPS2019_8358,massaroli2020stable}. \cite{hauser2019state} explored the algebraic structure of neural networks governed by finite difference equations, further linking discretizations of ODEs and ResNets in \citep{hauser2019state}.

Approximating ODEs with neural networks has been discussed in \citep{wang1998runge,filici2008neural}. \citep{poli2020hypersolvers} explores the interplay between Neural ODEs and their solver. On the optimization front, several works leverage dynamical system formalism in continuous time \citep{wibisono2016variational,maddison2018hamiltonian,massaroli2019port}.
\paragraph{Neural ODEs}
This work concerns Neural ODEs \citep{chen2018neural} and a system--theoretic discussion of their dynamical behavior. The main focus is on Neural ODEs and not the extensions to other classes of differential equations \citep{pmlr-v108-li20i,tzen2019neural,jia2019neural}, though the insights developed here can be broadly applied to continuous--depth models. More recently, \cite{finlay2020train} introduced regularization strategies to alleviate the heavy computational training overheads of Neural ODEs. These terms are propagated during the forward pass of the model and thus require state--augmentation. Leveraging our \textit{generalized adjoint} formulation provides an approach to integral regularization terms without augmentation and memory overheads.
%

%% file: 7_Discussion.tex
%
\vspace{-2mm}
\section{Conclusion}
In this work, we establish a general system--theoretic framework for Neural ODEs and dissect it into its core components. With the aim of shining light on fundamental questions regarding depth--variance, we formulate and solve the \textit{infinite--dimensional} problem linked to the true \textit{deep limit} formulation of Neural ODE. We provide numerical approximations to the infinite--dimensional problem, leading to novel model variants, such as Gal$\ddot{\text{e}}$rkin and piecewise--constant Neural ODEs. Augmentation is developed beyond existing approaches \citep{dupont2019augmented} to include \textit{input--layer} and \textit{higher--order} augmentation strategies showcased to be more performant and parameter efficient. Finally, the novel paradigms of data--control and depth--adaptation are introduced to perform challenging tasks such as learning \textit{reflections} without augmentation. The code to reproduce all the experiments present in the paper is built on {\tt TorchDyn} \citep{poli2020torchdyn} and {\tt PyTorch--Lighning} \citep{falcon2019pytorch} and can be found in the following repo: \url{https://github.com/DiffEqML/diffeqml-research/tree/master/dissecting-neural-odes}.  

%% file: Appendix_A.tex
\section{Proofs and Additional Theoretical Results}
\subsection{Proof of Theorem \ref{thm:GenAdj}}
\GenAdj*

\proof
\setlength{\abovedisplayskip}{2pt}
\setlength{\belowdisplayskip}{2pt}
Let us define a \textit{Lagrange multiplier} or \textit{adjoint} state $\ab$, dual to $\z$. As the dual of $\R^{\nz}$ is $\R^{\nz}$ itself, $\ab\in\R^{n_z}$. Moreover, let $\cL$ be a perturbed loss function of the form
    \[
        \cL := \ell - \int_0^{S} \ab^\top(\tau)\left[\dot \z(\tau) - f_{\theta}(s, \xb_t,\z(\tau))\right]\dd\tau
    \]
    Since $\dot \z - f_{\theta}(s, \xb, \z)=0$ by construction, the integral term in $\cL$ is always null and, thus, $\ab(s)$ can be freely assigned while {$\dd\cL/{\dd\theta} = \dd\ell/{\dd\theta}$}. For the sake of compactness we do not explicitly write the dependence on variables of the considered functions unless strictly necessary. Note that, 
    \begin{equation*}
        \begin{aligned}
            \int_0^{S} \ab^\top&\dot \z\dd\tau 
            = \ab^\top(\tau)\z(\tau)\big|_{0}^S - \int_0^{S} \dot\ab^\top \z\dd\tau
        \end{aligned}
    \end{equation*}
    obtained via integration by parts. Hence,
    \begin{equation}\label{eq:12}
        \begin{aligned}
            \cL &= \ell - \ab^\top(\tau)\z(\tau)\big|_{0}^S + \int_0^{S} \left(\dot\ab^\top \z + \ab^\top f_\theta \right)\dd\tau\\
            & = L(\z(S)) - \ab^\top(\tau)\z(\tau)\big|_{0}^S + \int_0^{S} \left(\dot\ab^\top \z + \ab^\top f_\theta + l \right)\dd\tau
        \end{aligned}
    \end{equation}
    We can compute the gradient of $\ell$ with respect to $\theta$ as 
    \begin{equation*}
        \begin{aligned}
            \frac{\dd\ell}{\dd\theta}& = \frac{\dd\cL}{\dd\theta} = \frac{\partial L(\z(S))}{\partial \z(S)}\frac{\dd \z(S)}{\dd \theta}- \ab^\top(S)\frac{\dd \z(S)}{\dd \theta}- \ab^\top(0)\cancel{\frac{\dd\z(0)}{\dd\theta}} \\
            & + \int_0^S \left[\dot\ab^\top\frac{\dd \z}{\dd\theta} +  \ab^\top\left(\frac{\partial f_\theta}{\partial \theta} + \frac{\partial f_\theta}{\partial \z}\frac{\dd\z}{\dd\theta} + \frac{\partial f_\theta}{\partial \xb}\cancel{\frac{\dd\xb}{\dd\theta}} + \frac{\partial f_\theta}{\partial \tau}\cancel{\frac{\dd\tau}{\dd\theta}} \right) + \frac{\partial l}{\partial\z}\frac{\dd\z}{\dd\theta} + \frac{\partial l}{\partial\tau}\cancel{\frac{\dd\tau}{\dd\theta}}\right]\dd\tau
        \end{aligned}
    \end{equation*}
    which, by reorganizing the terms, yields to 
    \begin{equation}\label{eq:fc_grad}
        \begin{aligned}
            \frac{\dd\ell}{\dd\theta} & = \left[\frac{\partial L}{\partial \z(S)} - \ab^\top(S)\right]\frac{\dd\z(S)}{\dd\theta} + \\
            & + \int_0^S \left(\dot\ab^\top + \ab^\top\frac{\partial f_\theta}{\partial \z} + \frac{\partial l}{\partial\z}\right)\frac{\dd \z}{\dd\theta}\dd\tau \\
            & + \int_0^S \ab^\top\frac{\partial f_\theta}{\partial \theta}\dd\tau
        \end{aligned}
    \end{equation}
    Now, if $\ab(s)$ satisfies the \textit{final} value problem
    \begin{equation}\label{eq:lm_ode}
        \begin{aligned}
            \dot\ab^\top(s) = -\ab^\top(s)\frac{\partial f_\theta}{\partial \z} - \frac{\partial l}{\partial\z},\quad\ab^\top(S) = \frac{\partial L}{\partial \z(S)}
        \end{aligned}
    \end{equation}
    to be solved backward in $[0,S]$; then (\ref{eq:fc_grad}) reduces to 
    \begin{equation}\label{eq:fc_grad_rd}
        \begin{aligned}
            \frac{\dd\ell}{\dd\theta} &= \int_0^S\ab^\top\frac{\partial f_\theta}{\partial \theta}\dd\tau
        \end{aligned}
    \end{equation}
    proving the result.
\endproof
\begin{rmk}[Implementation of the generalized adjoint method]
    Note that, similarly to \citep{chen2018neural}, the gradient (\ref{eq:fc_grad_rd}) is practically computed by defining the parameters adjoint state $\ab_\theta$ and solving backward the system of ODEs 
    \begin{equation}\label{eq:adj_ode}
        \begin{aligned}
            \dot\ab^\top &= -\ab^\top\frac{\partial f_\theta}{\partial \z}- \frac{\partial l}{\partial\z}, &&\ab^\top(S) = \frac{\partial L}{\partial \z(S)} \\
            \dot{\ab}_\theta^\top &= -\ab^\top\frac{\partial f_\theta}{\partial \theta}, &&\ab_\theta(S) = \mathbb{0}_{\nt}\\
        \end{aligned}
    \end{equation}
    Then,
    $$
        \frac{\dd\ell}{\dd\theta} = \ab_\theta(0).
    $$
\end{rmk}
\subsection{Proof of Theorem \ref{thm:2}}
\InfAdj*
\proof
The proof follows the same steps of the one of Theorem \ref{thm:GenAdj} up to \eqref{eq:12}. However, here $\theta(s)\in\mathbb{L}_2$ and the loss sensitivity to $\theta(s)$ corresponds to the directional (Gateaux) derivative $\delta\ell/\delta\theta(s)$ in $\bL_2$ derived as follows. We start by computing the total variation of $\ell$:
\begin{equation*}
    \begin{aligned}
        \delta\ell &= \frac{\partial L}{\partial \z(S)}\delta \z(S) - \ab^\top(s)(\delta \z(S) - \delta \z(0)) \\
        &+ \int_0^{S} \left[\dot{\ab}^\top(\tau) \delta \z(\tau) + \ab^\top(\tau) \left(\frac{\partial f_{\theta(\tau)}}{\partial \z(\tau)}\delta \z(\tau) + \frac{\partial f_{\theta(\tau)}}{\partial \theta(\tau)}\delta \theta(\tau)\right)+\frac{\partial l}{\partial \z(\tau)}\delta \z(\tau)\right]\dd\tau
    \end{aligned}
\end{equation*}
Thus,
\begin{equation*}
    \begin{aligned}
        &\frac{\delta\ell}{\delta \theta(s)} = \left[\frac{\partial L}{\partial \z(S)} - \ab^\top(s)\right]\frac{\delta \z(S)}{\delta \theta(s)}  + \frac{\delta\z(0)}{\delta \theta(s)}\\
        &~~+ \int_0^{S} \left[\dot{\ab}^\top(\tau) \frac{\delta \z(\tau)}{\delta \theta(s)}+ \ab^\top(\tau)\left(\frac{\partial f_{\theta(\tau)}}{\partial \z(\tau)}\frac{\delta \z(\tau)}{\delta \theta(s)} + \frac{\partial f_{\theta(\tau)}}{\partial \theta(\tau)}\frac{\delta \theta(\tau)}{\delta \theta(s)}\right) + \frac{\partial l}{\partial \z(\tau)}
        \frac{\delta \z(\tau)}{\delta\theta(s)}\right]\dd\tau
    \end{aligned}
\end{equation*}
Since it must hold
$$
    \int\frac{\delta\theta(\tau)}{\delta\theta(s)}\dd\tau = 1,
$$
then, model class choice $\theta(s)\in\bL_2$ implies
$$
    \frac{\delta\theta(\tau)}{\delta\theta(s)} = \delta(\tau - s)
$$
where $\delta(\tau-s)$ is the Dirac's delta. Therefore, it holds
\begin{equation*}
    \begin{aligned}
        &\frac{\delta\ell}{\delta \theta(s)} = \left[\frac{\partial L}{\partial \z(S)} - \ab^\top(s)\right]\frac{\delta \z(S)}{\delta \theta(s)}  + \frac{\delta\z(0)}{\delta \theta(s)}\\
        &~~+ \int_0^{S} \left[\dot{\ab}^\top(\tau) \frac{\delta \z(\tau)}{\delta \theta(s)}+ \ab^\top(\tau)\left(\frac{\partial f_{\theta(\tau)}}{\partial \z(\tau)}\frac{\delta \z(\tau)}{\delta \theta(s)} + \frac{\partial f_{\theta(\tau)}}{\partial \theta(\tau)}\delta(\tau - s)\right) + \frac{\partial l}{\partial \z(\tau)}
        \frac{\delta \z(\tau)}{\delta\theta(s)}\right]\dd\tau
    \end{aligned}
\end{equation*}
and, finally
\begin{equation*}
    \begin{aligned}
        &\frac{\delta\ell}{\delta \theta(s)} = \left[\frac{\partial L}{\partial \z(S)} - \ab^\top(s)\right]\frac{\delta \z(S)}{\delta \theta(s)}  + \frac{\delta\z(0)}{\delta \theta(s)}\\
        &~~+ \int_0^{S} \left(\dot{\ab}^\top(\tau) + \ab^\top(\tau)\frac{\partial f_{\theta(\tau)}}{\partial \z(\tau)} + \frac{\partial l}{\partial \z(\tau)}
       \right) \frac{\delta \z(\tau)}{\delta\theta(s)}\dd\tau\\
       &~~+\ab^\top(s)\frac{\partial f_{\theta(s)}}{\partial \theta(s)}
    \end{aligned}
\end{equation*}
Hence, if for any $s\in\cS$ the adjoint state $\ab(s)$ satisfies 
\begin{equation*}
    \begin{aligned}
        \dot\ab^\top &= -\ab^\top\frac{\partial f_{\theta(s)}}{\partial \z}- \frac{\partial l}{\partial\z}, &&\ab^\top(S) = \frac{\partial L}{\partial \z(S)} 
    \end{aligned}
\end{equation*}
we have 
$$ 
    \frac{\delta\ell}{\delta\theta(s)} = \ab^\top(s)\frac{\partial f_{\theta(s)}}{\partial \theta(s)}
$$
\endproof
\subsection{Proof of Corollary \ref{thm:cor1}}
\Cone*
\proof
The proof follows naturally from Theorem \ref{thm:2} by noticing that if $\theta(s)$ has some parametrization $\theta = \theta(s,\mu)$ with parameters $\mu\in\R^{n_\mu}$, then, 
\begin{equation}\label{eq:17}
    \frac{\dd \ell}{\dd \mu} = \int_0^S\ab^\top(\tau)\frac{\partial f_{\theta}}{\partial \theta}\frac{\partial \theta}{\partial\mu}\dd\tau
\end{equation}
Therefore, if
$$
    \theta(s) = \sum_{j=1}^m\alpha_j \odot \psi_j(s),
$$
the loss gradient with respect to the parameters $\alpha:=(\alpha_1,\dots,\alpha_m)\in\R^{m\nt}$ is computed as 
$$
    \begin{aligned}
        \frac{\dd \ell}{\dd \alpha} &= \int_0^S\ab^\top(\tau)\frac{\partial f_{\theta(\tau)}}{\partial \theta(\tau)}\frac{\partial \theta(s)}{\partial\alpha}\dd\tau \\
        &=\int_0^S\ab^\top(\tau)\frac{\partial f_{\theta(\tau)}}{\partial \theta(\tau)}\psi\dd\tau
    \end{aligned}
$$
being $\psi:=(\psi_1,\dots,\psi_m)$. 
\endproof
\begin{rmk}[Choose your parametrization]
    A further insight from this result, which paves the way to future developments, is that we can easily compute the loss gradients with respect to any parametrization of $\theta(s)$ through \eqref{eq:17}
\end{rmk}
%
\subsection{Proof of Corollary \ref{thm:cor2}}
\Ctwo*

\proof
    The proof follows from the one of Theorems \ref{thm:GenAdj} and \ref{thm:2} by recalling the solution of the stacked neural ODEs:
    $$ 
    \z(S) = h_x(\xb) + \sum_{i=0}^{p-1}\int_{s_1}^{s_{i+1}}f_{\theta_i}(\tau, \xb, \z(\tau))\dd \tau
    $$
    We can recover a relation similar to \eqref{eq:fc_grad} 
    \begin{equation*}
        \begin{aligned}
            \frac{\dd\ell}{\dd\theta_i} & = \left[\frac{\partial L}{\partial \z(S)} - \ab^\top(S)\right]\frac{\dd\z(S)}{\dd\theta_i} + \\
            & + \sum_{j=0}^{p-1}\int_{s_j}^{s_{j+1}} \left(\dot\ab^\top + \ab^\top\frac{\partial f_{\theta_j}}{\partial \z} + \frac{\partial l}{\partial\z}\right)\frac{\dd \z}{\dd\theta_i}\dd\tau \\
            & + \sum_{j=0}^{p-1}\int_{s_j}^{s_{j+1}} \ab^\top\frac{\partial f_{\theta_j}}{\partial \theta_i}\dd\tau
        \end{aligned}
    \end{equation*}
    Since 
    $$
        \forall j = 0,\dots, p-1~~~~\frac{\partial f_{\theta_j}}{\partial\theta_i} \neq \mathbb{0} \Leftrightarrow j = i,
    $$
    we have
    $$
        \sum_{j=0}^{p-1}\int_{s_j}^{s_{j+1}} \ab^\top\frac{\partial f_{\theta_j}}{\partial \theta_i}\dd\tau = \int_{s_i}^{s_{i+1}} \ab^\top\frac{\partial f_{\theta_i}}{\partial \theta_i}\dd\tau =  -\int_{s_{i+1}}^{s_{i}} \ab^\top\frac{\partial f_{\theta_i}}{\partial \theta_i}\dd\tau
    $$
    which leads to the result by assuming $\ab(\tau)$ to satisfy
    $$
        \begin{aligned}
            \dot{\ab}^\top(s) &= -\ab^\top(s)\frac{\partial f_{\theta_i}}{\partial \z} - \frac{\partial l}{\partial \z}~~~s\in[s_i,s_{i+1}]\\
            \ab^\top(S) &=  \frac{\partial L}{\partial \z(S)}
        \end{aligned}
    $$ 
\endproof

\subsection{Proof of Theorem \ref{th:san}}
\san*
\proof\label{proof}
    The general solution of (\ref{eq:lin_sys}) is
    \[
        z(s) = x(2e^{-\theta s} - 1)
    \]
    Thus,
    \[
        \begin{aligned}
            &e = \varphi(x) - z(1) = x + x(2e^{-\theta}-1) = 2xe^{-\theta}\\
            \Leftrightarrow ~& |e| = 2|x|e^{-\theta}
        \end{aligned}
    \]
    It follows that 
    \begin{align*}
        & 2|x|e^{-\theta}<\epsilon\\
        \Leftrightarrow ~& e^{-\theta}<\frac{\epsilon}{2|x|}\\
        \Leftrightarrow ~& \theta>-\ln\left(\frac{\epsilon}{2|x|}\right)
    \end{align*}
\endproof
\subsection{Additional Theoretical Results}
\subsubsection{Explicit Parameter Dependence of the Loss}
Note that, in both the seminal paper from \cite{chen2018neural} and Theorem \ref{thm:GenAdj} the loss function was consider without explicit dependence on the parameters. However, in practical applications (see, e.g. \citep{finlay2020train}) the loss has this explicit dependence: 
\begin{equation}\label{eq:18}
        \ell = L(\z(S), \theta) + \int_\cS l(s,\z(\tau), \theta)\dd \tau,
\end{equation}
In this case we need to modify the adjoint gradients accordingly
\begin{theorem}[Generalized Adjoint Method with  Parameter--Dependent Loss]
Consider the loss function \eqref{eq:18}. Then,
$$
    \frac{\dd \ell}{\dd \theta} = \frac{\partial L}{\partial\theta} + \int_{\cS}\left(\ab^\top(\tau)\frac{\partial f_\theta}{\partial\theta} + \frac{\partial l}{\partial\theta}\right)\dd \tau
$$
where $\ab(s)$ satifies \eqref{eq:lm_ode}.
\end{theorem}
\proof
The proof follows immediately from Theorem \ref{thm:GenAdj} by noticing that, with the explicit dependence on $\theta$ of $\ell$, \eqref{eq:fc_grad} would become
\begin{equation*}
        \begin{aligned}
            \frac{\dd\ell}{\dd\theta} & = \frac{\partial L}{\partial\theta}\\
            &+\left[\frac{\partial L}{\partial \z(S)} - \ab^\top(S)\right]\frac{\dd\z(S)}{\dd\theta} + \\
            & + \int_0^S \left(\dot\ab^\top + \ab^\top\frac{\partial f_\theta}{\partial \z} + \frac{\partial l}{\partial\z}\right)\frac{\dd \z}{\dd\theta}\dd\tau \\
            & + \int_0^S \left(\ab^\top\frac{\partial f_\theta}{\partial \theta} + \frac{\partial l}{\partial \theta}\right)\dd\tau
        \end{aligned}
    \end{equation*}
leading to the result.
\endproof
In the depth--variant case where we might consider a loss function of type
\begin{equation}
    \ell = L(\z(S), \theta(S)) + \int_\cS l(\z(\tau), \theta(\tau))\dd \tau
\end{equation}
a similar result can be obtained for the infinite--dimensional adjoint.
\subsubsection{Integration Bound Gradients}
It is also possible to obtain the loss gradient with respect to the integration bound $S$.
\begin{theorem}[Integration Bound Gradient]
    Consider a loss function \ref{eq:2}. Then, 
    $$
        \frac{\dd \ell}{\dd S} = \frac{\partial L}{\partial \z(S)}f_{\theta(S)}(S, \xb, \z(S)) + l(\z(S))
    $$
\end{theorem}
\proof
    \begin{equation*}
        \begin{aligned}
        \frac{\dd\ell}{\dd S} &= \frac{\partial L}{\partial \z(S)}\frac{\dd \z(S)}{\dd S} + \frac{\dd }{\dd S}\int_0^S l(\z(\tau))\dd\tau\\
        & = \frac{\partial L}{\partial \z(S)}\frac{\dd }{\dd S}\left(h_x(\xb) + \int_0^Sf_{\theta(\tau)}(\tau, \xb, \z(\tau))\right) + \frac{\dd }{\dd S}\int_0^S l(\z(\tau))\dd\tau
        \end{aligned}
    \end{equation*}
    Therefore, by applying the Leibniz integral rule we obtain
    $$
        \frac{\dd \ell}{\dd S} = \frac{\partial L}{\partial \z(S)}f_{\theta(S)}(S, \xb, \z(S)) + l(\z(S))
    $$
\endproof

%% file: Appendix_B.tex
\section{Practical Insights for Neural ODEs}
\subsection{Augmentation}
\paragraph{Augmenting convolution and graph based architectures}
In the case of \textit{convolutional neural network} (CNN) or \textit{graph neural network} (GNN) architectures, augmentation can be performed along different dimensions i.e. \textit{channel}, heigth, width or similarly node features or number of nodes. The most physically consistent approach, employed in \citep{dupont2019augmented} for CNNs, is augmenting along the \textit{channel} dimension, equivalent to providing each pixel in the image additional states. By viewing an image as a lattice graph, the generalization to GNN--based Neural ODEs \citep{poli2019graph} operating on arbitrary graphs can be achieved by augmenting each node feature with $n_a$ additional states. 

\paragraph{Selective higher--order}
A limitation of system (\ref{eq:HO_aug}) is that a naive extension to second--order requires a number of augmented dimensions $n_a = n_z/2$. To allow for flexible augmentations of few dimensions $n_a <n_z/2$, the formulation of second--order Neural ODEs can be modified as follows. Let $\z:=(\z_q,\z_p,\bar\z)$, $\z_q,\z_p\in\R^{n_a/2},~\bar\z\in\R^{n_z-n_a}$.
We can decide to give second order dynamics only to the first $n_a$ states while the dynamics of other $n_z-n_a$ states is free. Therefore, this approach yields
\begin{equation}
    \begin{aligned}
        \begin{bmatrix}\dot\z_q\\\dot\z_p\\\dot{\bar\z}\end{bmatrix} &= \begin{bmatrix}\z_p\\ f^p_{\theta(s)}(s, \z)\\ \bar f_{\theta(s)}(s, \z)\end{bmatrix} 
    \end{aligned},
\end{equation}
A similar argument could be applied to orders higher than two. \textit{Selective higher--order} Neural ODEs are compatible with input layer augmentation.
\begin{figure*}[b]
    \centering
    \includegraphics[width=.75\linewidth]{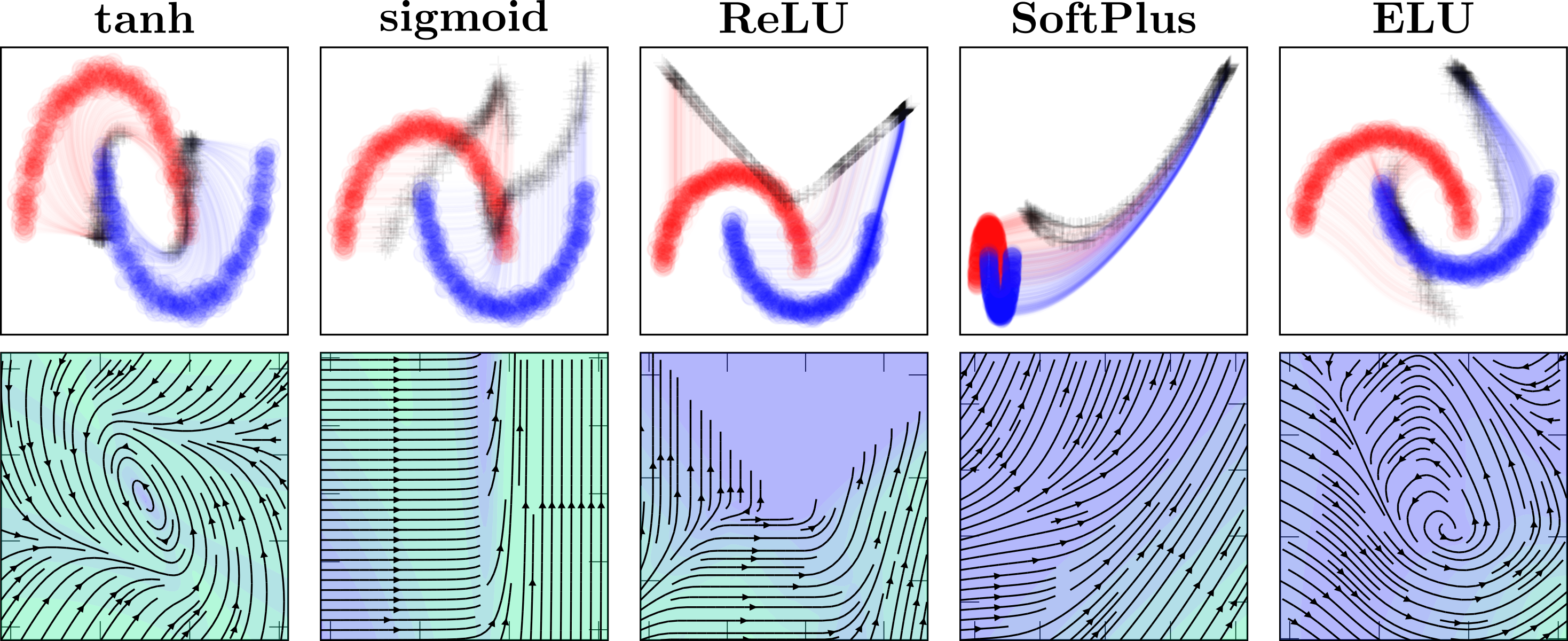}
    \vspace*{-3mm}
    \caption{Depth trajectories of the hidden state and relative vector fields $f_\theta(\z)$ for different activation functions in a nonlinear classification task. It can be noticed how the models with tanh and ELU outperform the others, as $f_\theta$ is able to steer $\z$ along negative directions.}
    \label{fig:activations}
\end{figure*}
\subsection{Activations}
\paragraph{Mind your activation} We investigate the effects of appending an activation function to the last layer of $f_{\theta}$. The chosen nonlinearity will strongly affect the ``shape'' of the vector field and, as a consequence, the flows learnable by the model. Therefore, while designing $f_{\theta}$ as a multi--layer neural network, it is generally advisable to append a linear layer to maximize the expressiveness of the underlying vector field. In some applications, conditioning the vector field (and thus the flows) with a specific nonlinearities can be desirable, e.g., when there exist priors on the desired transformation, such as boundedness of the vector field.
\paragraph{Effects of activations} In order to compare the effect of different activation functions in the last layer of $f_{\theta}$, we set up a nonlinear classification task with the half--moons dataset. For the sake of completeness, we selected activations of different types, i.e., 
\begin{table}[h]
    \centering
    \begin{tabular}{rc}
    \toprule
    Activation & Type\\
    \midrule
    Hyperbolic tangent (\textbf{tanh}) & bounded\\
    \textbf{Sigmoid}&  bounded, non--negative output\\
    \textbf{ReLU}& unbounded, non--negative output\\
    \textbf{Softplus}& unbounded, non--negative output\\
    \textbf{ELU}& lower--bounded\\
    \bottomrule
    \end{tabular}
\end{table}
The dataset is comprised of $2^{13}$ data points. We utilize the entire dataset for training and evaluation since the experiment has the aim of delivering a qualitative description of the learned vector fields.  $f_\theta$ has been selected as a multilayer perceptron with two hidden layers of 16 neurons each. The training has been carried out using Adam \citep{kingma2014adam} optimizer with learning rate $10^{-3}$ and weight decay set to $10^{-4}$.

In Figure \ref{fig:activations} we see how different activation functions in the last layer of $f_\theta$ condition the vector fields and the depth evolution of the hidden state in the classification of nonlinearly separable data. It is worth to be noticed that the models with better performance are the ones with \textit{hyperbolic tangent} (tanh) and ELU \citep{clevert2015fast} as the vector field can assume both positive and negative values and, thus, can ``force'' the hidden state in different directions. On the other hand, with sigmoid, ReLU or softplus \citep{zheng2015improving}, the vector field is nonnegative in all directions and thus has limited freedom. Further, Figure \ref{fig:activations_db} shows how different activation functions shape the vector field and as a result the decision boundary.
\subsection{Regularization for Stability}
The concept of stability can be used to regularize Neural ODEs through a variety of additional terms or different formulations \citep{finlay2020train, massaroli2020stable}. \citep{finlay2020train} proposes minimizing a loss term:
\begin{equation}\label{tisode}
\ell_{\tt reg} = \int_\cS\|f_{\theta(\tau)}(\tau,\xb,\z(\tau)\|_2\dd\tau,
\end{equation}
to achieve stability. A simple alternative stabilizing regularization term can be considered at no significant additional computational cost:
\begin{equation}\label{new_tis}
\ell_{\tt reg} = \norm{f_{\theta(S)}(S, \xb, \z(S))}_2,
\end{equation}
which penalizes non--convergence to some fixed point of $f_\theta$ at $s=S$. The above can also be seen as a cheaper alternative to the kinetic energy regularization proposed in \citep{finlay2020train}.
\subsection{Approximation Capabilities}
Vanilla Neural ODEs are not, in general, universal function approximators (UFAs) \citep{zhang2019approximation}. Besides some recent works on the topic \citep{zhang2019approximation, li2019deep} this apparent limitation is still not well--understood in the context of continuous--depth models. When Neural ODEs are employed as general--purpose black--box modules, some assurances on the approximation capabilities of the model are necessary. Let $\nz := \nx + 1$ and let $\z:=(\z_x, z_a)$ ($\z_x\in\R^\nx,~~z_a\in\R$). \citep{zhang2019approximation} noticed that a depth--invariant augmented Neural ODE
\begin{equation}
    \begin{matrix*}[l]
    \begin{bmatrix}
        \dot\z_x\\
        \dot z_a
    \end{bmatrix} =  
    \begin{bmatrix}
        \mathbb{0}_\nx\\
        f_{\theta}(\z_x)
    \end{bmatrix},~~
    \begin{bmatrix}
        \z_x(0)\\
        z_a(0) 
    \end{bmatrix} = 
    \begin{bmatrix} 
        \xb\\
        0
    \end{bmatrix}
    \end{matrix*},~~s\in[0,1]
\end{equation}
where the output is picked as $\hat y := z_a(1)$, can approximate any function $\Psi:\R^\nx\rightarrow\R$ provided that the neural network $f_\theta(\xb)$ is an approximator of $\Psi$, since $z_a(1) = f_\theta(\xb)$, mimicking the mapping $\xb\mapsto f_\theta(\xb)$. Although this simple result is not sufficient to provide a constructive blueprint to the design of Neural ODE models, it suggests the following (open) questions:
\begin{itemize}
    \item Why should we use a Neural ODE if its vector field can solve the approximation problem as a standalone neural network?
    \item Can Neural ODEs be UFAs with non-UFA vector fields?
\end{itemize}
On the other hand, if Neural ODEs are used for model discovery or observation of dynamical systems, requiring an UFA neural network to parametrize the model provides it with the ability to approximate arbitrary dynamical systems.
\subsection{Example Implementation of Data--Control}
We report here a short PyTorch code snippet detailing the implementation of the simplest \textit{data--controlled Neural ODE} variant, accompanied, for further accessibility, by a brief text description.

\begin{minted}{Python}
class DC_DEFunc(nn.Module):
    """PyTorch implementation of data--controlled $f_\theta$"""
    def __init__(self, f):
        super().__init__()  
        self.f = f
        
    def forward(self, s, z):
    """Forward is called by the ODE solver repeatedly"""
        self.nfe += 1
        # data-control step: 
        # alternatives include embeddings of input data `x` i.e g(x)
        # or addition `x + z`
        z = torch.cat([z, self.x], 1) 
        dz = self.f(z)
        return dz
\end{minted}

where the initial condition $\xb$ is passed to the model at the start of the integration at $s=0$. The information contained is thus passed repeatedly to the function $f_{\theta}$, conditioning the dynamics. It should be noted that even in the case of concatenation of $\xb$ and $\z(s)$, the above is not a form of augmentation, since the state itself is not given additional dimensions during forward propagation. In fact, the dynamics take the form of a function $f_{\theta}:\R^{n_x}\times\R^{n_z}\rightarrow\R^{n_z}$ instead of $f_{\theta}:\R^{n_a}\times\R^{n_x}\rightarrow\R^{n_a}\times\R^{n_x}$ as is the case for general first--order augmentation with $n_z = n_x + n_a$.

%% file: Appendix_C.tex
\section{Experimental Details}
\paragraph{Computational resources}
The experiments were carried out on a cluster of two \textsc{NVIDIA\textsuperscript{\textregistered} Titan RTX} GPUs with CUDA 10.1 and \textsc{Intel\textsuperscript{\textregistered} i9 10980xe} CPU. All Neural ODEs were trained on GPU. The code was built upon Pytorch's {\tt torchdyn} library for neural differential equations \citep{poli2020torchdyn}.
\paragraph{General experimental setup}\label{par:gen}
We report here general information about the experiments. All Neural ODEs are solved numerically via the \textit{Dormand--Prince} method \citep{prince1981high}. We refer to \textit{concat} as the depth--variant Neural ODE variants where the depth--variable $s$ is concatenated to $\z(s)$ as done in \citep{chen2018neural}. Furthermore, we denote Gal\"erkin Neural ODEs as \textit{GalNODE} for convenience.
\paragraph{Benchmark problems}
Throughout the paper we extensively utilize the \textit{concentric annuli} benchmark task introduced in \citep{dupont2019augmented} is used extensively. Namely, given $r>0$ define $\varphi: \R^n \to \mathbb{Z}$
\begin{equation}\label{benchmark}
    \varphi(\xb) =
    \left\{
        \begin{matrix*}[l]
            -1 & \norm{\xb}_2 < r\\%
             1 & \norm{\xb}_2 \geq r%
        \end{matrix*}
    \right.~.
\end{equation}
We consider learning the map $\varphi(\xb)$ with Neural ODEs prepending a linear layer $\R^n \rightarrow \R$. Notice that $\varphi$ has been slightly modified with respect to \citep{dupont2019augmented}, to be \textit{well--defined} in its domain. For the one--dimensional case, we will often instead refer to the map $\varphi(x) = -x$ as the \textit{crossing trajectories} problem. The optimization is carried out by minimizing \textit{mean squared error} (MSE) losses of model outputs and mapping $\varphi$.
\begin{figure*}[t]
    \centering
    \includegraphics[scale =.9]{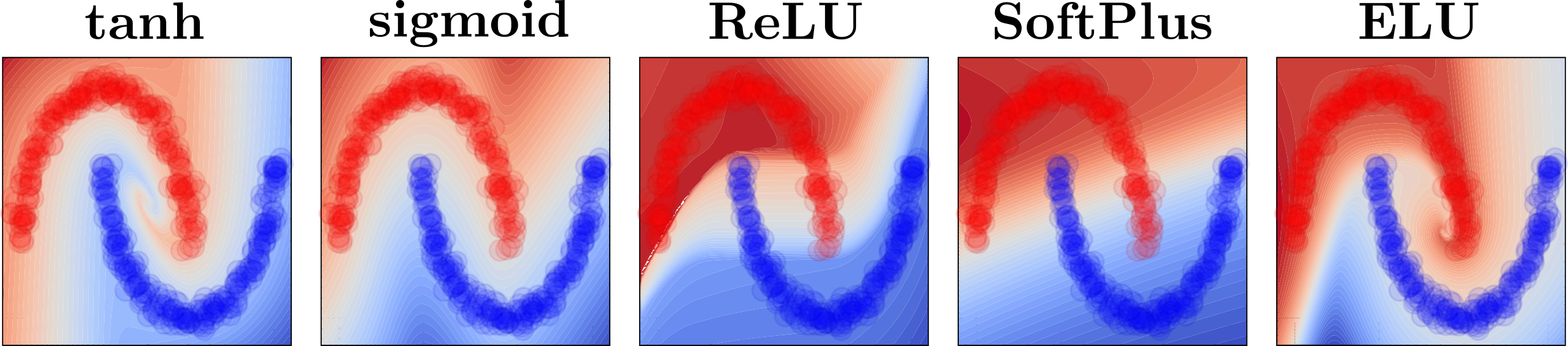}
    \vspace{-1mm}
    \caption{Decision boundaries learned by the vector field of a Neural ODE are directly conditioned by the choice of activation function.}
    \label{fig:activations_db}
\end{figure*}
\subsection{Experiments of Section \ref{sec:dissecting}}
\paragraph{Trajectory tracking}
Consider the problem of tracking a periodic signal $\beta(s)$. We show how this can be achieved without introducing additional inductive biases such as \citep{greydanus2019hamiltonian} through a synergistic combination of a two--layer Gal\"erkin Neural ODEs and the generalized adjoint with integral loss $l(s):=\|\beta(s)-\z(s)\|^2_2$. In particular, we construct a two--layer Gal\"erkin Neural ODE with Fourier series and $m = 2$ harmonics as the eigenfunctions. The training is carried out for $1000$ epochs with learning rate $10^{-3}$. The practical implementation of the generalized adjoint necessary to distribute the loss across the depth domain is discussed in Appendix A.

The models, trained in $s\in[0,1]$ generalize accurately when tasked to perform long trajectory extrapolation of several seconds. 
\paragraph{Depth--varying classification}
We showcase how different discretization options of the functional optimization problem discussed in Sec.~\ref{sec:dissecting} affect the final dynamics of $\theta(s)$. Namely, we consider a simple binary classification on the \textit{nested spirals} problem, training all models for $300$ epochs and learning rate $5 \cdot 10^{-3}$. Gal\"erking Neural ODEs are equipped with a polynomial basis with $m = 10$. The Fig.s in Sec~\ref{sec:dissecting} reveal the different nature of $\theta(s)$ depending on model choice: depth--discretization of \textit{Stacked} yields a flexible, though lower resolution form of $\theta(s)$, whereas spectral discretizations limit the functional form of $\theta(s)$ to the span of a chosen eigenbasis.
\paragraph{Mind your input network experiments}
We tackle the \textit{concentric annuli} task with a Neural ODE preceded by a simple two--layer neural network with $16$ units and ReLU activation. The second layer is linear. 
\subsection{Experiments of Section \ref{sec:augmenting}}
\paragraph{Image classification}
We use \textit{AdamW} with learning rate $10^{-3}$, batch size $64$, weight decay $5*10^{-4}$ and a learning rate step schedule with multiplicative factor $\gamma=0.9$ every $5$ epochs. We train each model for $20$ epochs. The vector fields $f_\theta$ are parametrized by 3--layer depth--invariant CNNs, with each layer followed by an instance normalization layer. The choice of depth--invariance is motivated by the discussion carried out in Section \ref{sec:no_aug}: both augmentation and depth--variance can relieve approximation limitations of vanilla, depth--invariant Neural ODEs. As a result, including both renders the ablation study for augmentation strategies less accurate. We note that the results of this ablation analysis do not utilize any form of data augmentation; data augmentation can indeed be introduced to further improve performance.

For input layer augmented Neural ODE models, namely IL--NODE and 2nd order, we prepend to the Neural ODE a single, linear CNN layer. In the case of 2nd order models, we use input layer augmentation for the positions and initialize the velocities at $0$. The hidden channel dimension of the CNN parametrizing $f_\theta$ in augmented models is set to $32$ on MNIST and $42$ on CIFAR; vanilla Neural ODEs, on the other hand, are equipped with dimensions $42$ and $62$ for a fair comparison. The output class probabilities are then computed by mapping the output of the Neural ODE through average pooling followed by a linear layer. Second order Neural ODEs, \textit{2nd}, use $f_\theta$ to compute the vector field of velocities: therefore, the output of $f_\theta$ is ${\nx}/{2}$--dimensional, and the remaining ${\nx}/{2}$ outputs to concatenate (vector field of \textit{positions}) are obtained as the last ${n_x}/{2}$ elements of $\z$. 

We note that vanilla Neural ODEs are capable of convergence without any spikes in loss or NFEs. We speculate the numerical issues encountered in \citep{dupont2019augmented} to be a consequence of the specific neural network architecture used to parametrize the vector field $f_{\theta}$, which employed an excessive number of channels inside $f_\theta$, i.e $92$.
\subsection{Experiments of Section \ref{sec:no_aug}}
\begin{figure*}
        \centering
        \includegraphics{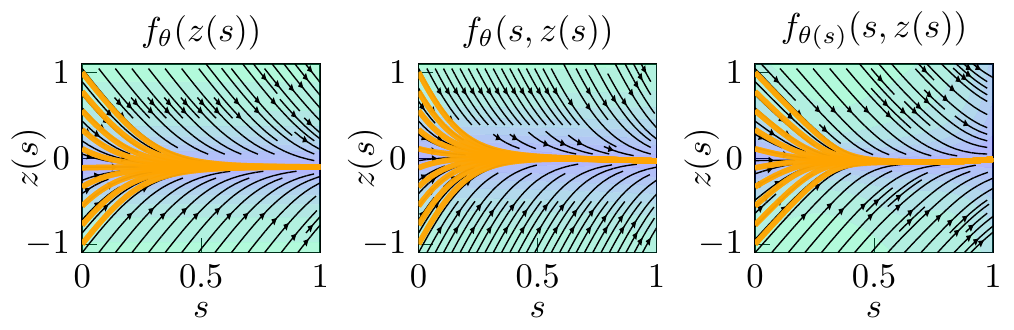}
        \vspace{-5mm}
        \caption{Depth evolution over the learned vector fields of the standard models: depth--invariant and depth--variant (``concat'' $f_\theta(s,z(s))$ and GalNODE $f_{\theta}(s, z(s))$). As expected the Neural ODE cannot approximate the map $\varphi(x) = -x$.}
        \label{fig:1d_vanilla}
\end{figure*}

\begin{figure}
    \centering
    \includegraphics[width=.9\linewidth]{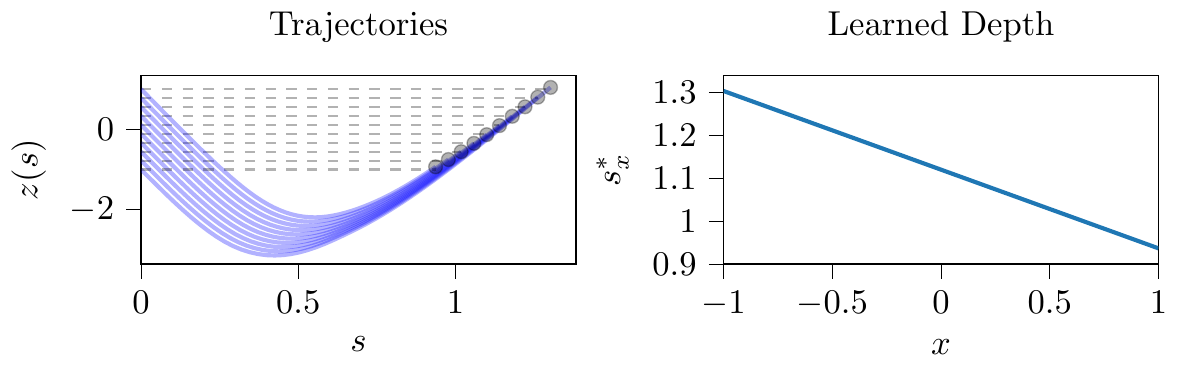}
    \vspace{-3mm}
    \caption{Evolution of the input data through the depth of the Neural ODEs}
    \label{fig:adapt_time2}
\end{figure}

\paragraph{Experiments on crossing trajectories}
We trained both current state--of--the--art as well as proposed models to learn the map $\varphi(x) = -x$. We created a training dataset sampling $x$ equally spaced in $[-1,1]$. The models have been trained to minimize L1 losses using Adam \citep{kingma2014adam} with learning rate $lr = 10^{-3}$ and weight decay $10^{-5}$ for 1000 epochs using the whole batch. We trained vanilla Neural ODEs, i.e. both depth--invariant and depth variant models (``concat'' and GalNODE). As expected, these models cannot approximate $\varphi$. Both depth--invariant and \textit{concat} have been selected with two hidden layers of 16 and 32 neurons each, respectively and \textit{tanh} activation. The GalNODE have been designed with one hidden layer of 32 neurons whose depth--varying weights were parametrized by a Fourier series of five modes. The resulting trajectories over the learned vector fields are shown in Fig. \ref{fig:1d_vanilla}.
\paragraph{Data--controlled Neural ODEs} We evaluate both the handcrafted linear depth--invariant model (\ref{eq:lin_sys}) and the general formulation of data--controlled models (\ref{eq:cnode}), realized with two hidden layers of 32 neurons each and \textit{tanh} activation in all layers but the output. Note that the loss of the handcrafted model results to be convex and continuously differentiable. Moreover, proof \ref{proof} provides analytically a lower bound on the model parameter to ensure the loss to be upper--bounded by a desired $\epsilon$, making its training superfluous. Nevertheless, we provide results with a trained version to show that the benefits of data--controlled Neural ODEs are compatible with gradient--based learning. 
\begin{figure*}
    \centering
    \includegraphics{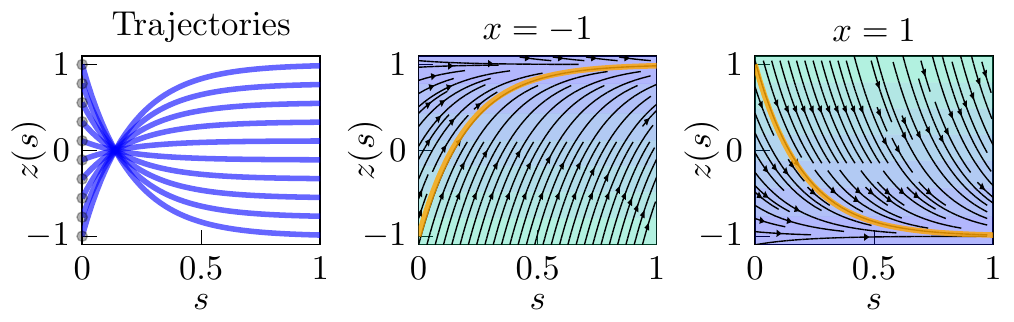}
    \includegraphics{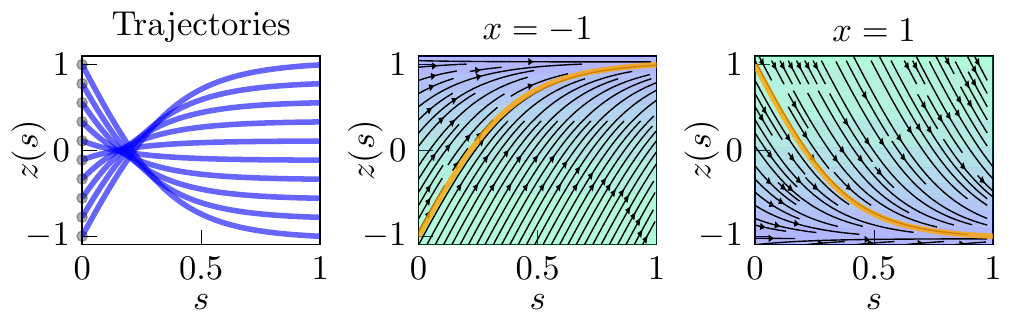}
    \vspace{-3mm}
    \caption{Depth evolution over the learned vector fields of \eqref{eq:lin_sys} and a data--controlled Neural ODE. As discussed in Sec.\ref{sec:no_aug} introducing data--control allows the model to approximate the map $\varphi(x) = -x$.}
    \label{fig:1d_controlled}
\end{figure*}
The results are shown in Fig.s \ref{fig:1d_vanilla} and \ref{fig:1d_controlled}. The input data information embedded into the vector field allows the Neural ODE to steer the hidden state towards the desired label through its continuous depth. Data--controlled Neural ODEs can be used to learn challenging maps \citep{dupont2019augmented} without augmentation.
\paragraph{Concentric annuli with non--augmented variants}
We train each model for $1024$ iterations using AdamW with learning rate $10^{-3}$, weight decay $10^{-6}$ and batch size $1024$. All models have a single hidden layer of dimension $32$. The GalNODE layer is parametrized by a Fourier series of five modes.
\paragraph{Conditional continuous normalizing flows}
We train data--controlled continuous normalizing flows for $2000$ iterations with samples of size $2^{14}$. We use AdamW with learning rate $10^{-3}$ and weight decay $10^{-7}$. Absolute and relative tolerances of the chosen solver, {\tt dopri5} are set to $10^{-8}$. The CNF network have $2$ hidden layers of dimension $128$ with softplus nonlinearities. 
\paragraph{Adaptive depth Neural ODEs}
The experiments have been carried out with a depth--variant Neural ODE in ``concat'' style where $f$ was parametrized by a neural network with two hidden layers of 8 units and $\tanh$ activation. Moreover, the function $g_{\omega}(\xb)$ computing the data--adaptive depth of the Neural ODE was composed by a neural network 
with one hidden layer (8 neurons and ReLU activation) whose output is summed to one and then taken in absolute value,   
\[ 
    g(\xb) = \left|1 + \wb_o^\top\sigma(\wb_i\xb + \mathbf{b}_i) + b_o\right|
\]
where $\sigma$ is the ReLU activation, $\wb_o,~\wb_i,~\mathbf{b}_i\in\R^{8}$ and $\omega = (\wb_o, b_o, \wb_i, \mathbf{b}_i)$. In particular, the summation to one has been employed to help the network ``sparsify'' the learned integration depths and avoid highly stiff vector fields, while the absolute value is needed to avoid infeasible integration intervals. The training results can be visualized in Fig. \ref{fig:adapt_time2}. This early result should be intended as a proof of concept rather than a definitive evaluation of the depth adaptation methods, which we reserve for future work. We note that the result of Fig. \ref{fig:adapt_time} showed in the main text has been obtained by training the model only on $x\in\{-1,1\}$ and manually setting $s_{-1}^* = 1,~s_{1}^* = 3$.
